\documentclass[letterpaper]{article} % DO NOT CHANGE THIS
\usepackage{aaai25}  % DO NOT CHANGE THIS
\usepackage{times}  % DO NOT CHANGE THIS
\usepackage{helvet}  % DO NOT CHANGE THIS
\usepackage{courier}  % DO NOT CHANGE THIS
\usepackage[hyphens]{url}  % DO NOT CHANGE THIS
\usepackage{graphicx} % DO NOT CHANGE THIS
\usepackage{natbib}  % DO NOT CHANGE THIS AND DO NOT ADD ANY OPTIONS TO IT
\usepackage{caption} % DO NOT CHANGE THIS AND DO NOT ADD ANY OPTIONS TO IT
\usepackage{algorithm,amsmath,amsfonts}
\usepackage{algorithmic,algorithmic}
\usepackage{float,multirow,booktabs,cite} 
\usepackage[T1]{fontenc} 
\usepackage{pifont}
\usepackage{xcolor}

\definecolor{linkcolor}{rgb}{0, 0, 1} % 定义蓝色
\usepackage{newfloat}
\usepackage{listings}
\DeclareCaptionStyle{ruled}{labelfont=normalfont,labelsep=colon,strut=off} % DO NOT CHANGE THIS
\floatstyle{ruled}
\newfloat{listing}{tb}{lst}{}
\floatname{listing}{Listing}
\nocopyright 
\definecolor{darkgreen}{rgb}{0.0, 0.8, 0.0}
\definecolor{darkred}{rgb}{0.8, 0.0, 0.0}
\definecolor{customPink}{RGB}{255,192,203}
\title{FreeMask: Rethinking the Importance of Attention Masks \\ for Zero-Shot Video Editing}

\author{
    Lingling Cai\textsuperscript{\rm 1},
    Kang Zhao\textsuperscript{\rm 2},
    Hangjie Yuan\textsuperscript{\rm 2,1},
    Yingya Zhang\textsuperscript{\rm 2},
    Shiwei Zhang\textsuperscript{\rm 2},
    Kejie Huang\textsuperscript{\rm 1}\thanks{Corresponding author}
}
\affiliations{
    \textsuperscript{\rm 1}Zhejiang University,
    \textsuperscript{\rm 2}Alibaba Group
    \vspace{.2cm} \\
    \texttt{Project page: \color{customPink}\url{https://freemask-edit.github.io/}}
}

\usepackage{bibentry}
\begin{document}

\maketitle

\begin{abstract}
Text-to-video diffusion models have made remarkable advancements. Driven by their ability to generate temporally coherent videos, research on zero-shot video editing using these fundamental models has expanded rapidly.
To enhance editing quality, structural controls are frequently employed in video editing. Among these techniques, cross-attention mask control stands out for its effectiveness and efficiency.
However, when cross-attention masks are naively applied to video editing, they can introduce artifacts such as blurring and flickering.
Our experiments uncover a critical factor overlooked in previous video editing research: cross-attention masks are not consistently clear but vary with model structure and denoising timestep. 
To address this issue, we propose the metric Mask Matching Cost (MMC) that quantifies this variability and propose \textbf{FreeMask}, a method for selecting optimal masks tailored to specific video editing tasks.
Using MMC-selected masks, we further improve the masked fusion mechanism within comprehensive attention features, e.g., temp, cross, and self-attention modules.
Our approach can be seamlessly integrated into existing zero-shot video editing frameworks with better performance, requiring no control assistance or parameter fine-tuning but enabling adaptive decoupling of unedited semantic layouts with mask precision control. 
Extensive experiments demonstrate that FreeMask achieves superior semantic fidelity, temporal consistency, and editing quality compared to state-of-the-art methods.

\end{abstract}


\section{\bf{Introduction}}
With the growing interest in video diffusion models~\cite{wang2023modelscopet2v,yuan2024instructvideo,ho2022imagen}, research on video editing based on these foundational models is booming, which aims at high-performance video re-creation.
Among these editing paradigms, zero-shot video editing has attracted considerable attention due to its efficiency and low computational cost. 
Recent zero-shot video editing approaches~\cite{bai2024uniedit,ku2024anyv2v} have leveraged open-source pre-trained text-to-video diffusion models~\cite{wang2023lavie,zhang2023i2vgen} to improve temporal consistency, effectively alleviating issues like visual flickering across frames, which often occurred in earlier image-based methods~\cite{qi2023fatezero,geyer2023tokenflow}.
\begin{figure}[t]  
  \centering 
  \includegraphics[width=0.5\textwidth]{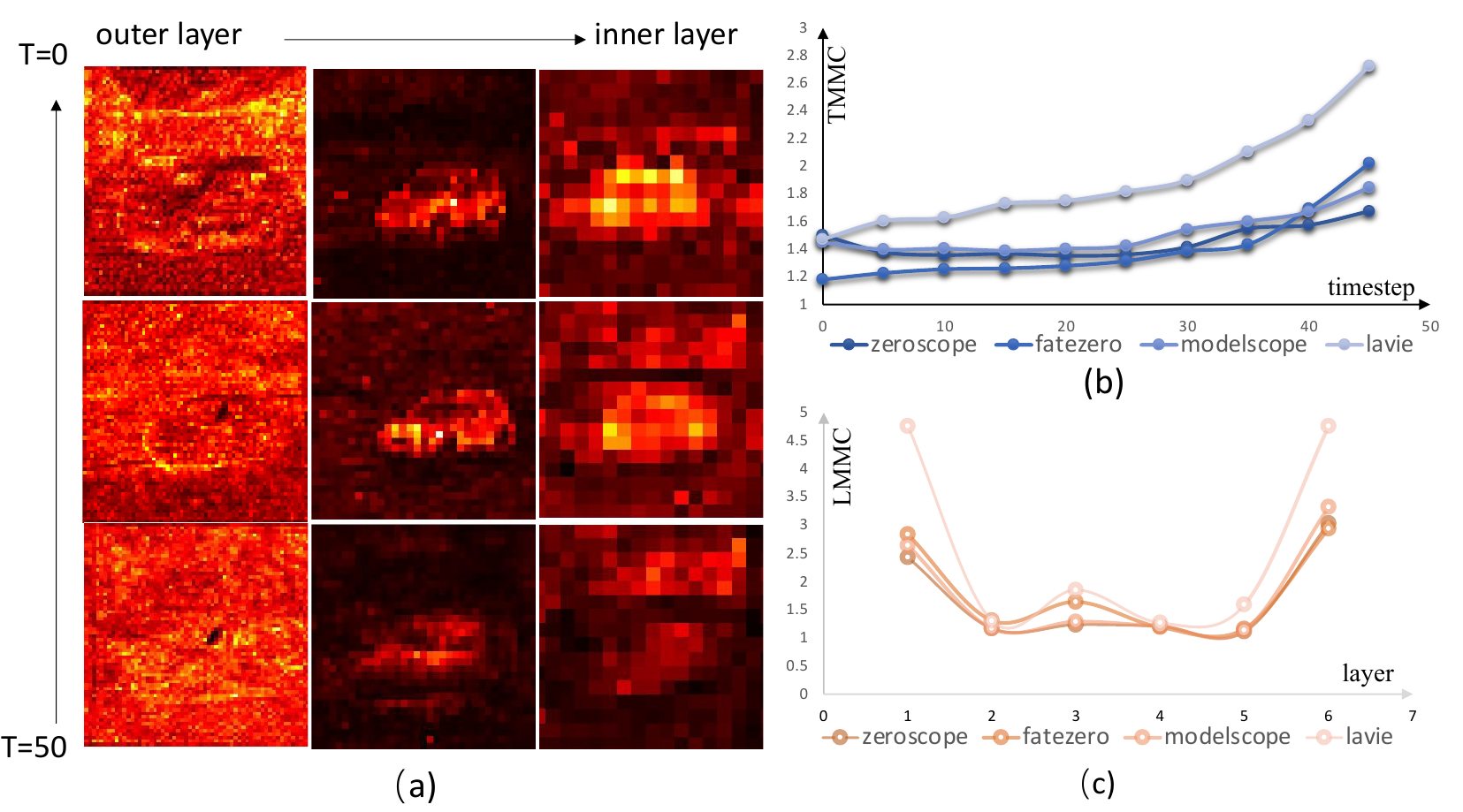} 
  \vspace{-.7cm}
  \caption{
    (a) Visualization of the 'jeep' cross-attention maps across layers and denoising timesteps on zeroscope~\cite{zeroscopev2}. (b) TMMC of different models across timesteps. (c) LMMC of different models across layers.}
  \vspace{-.3cm}
  \label{fig:insight1}   
\end{figure}

To further improve editing quality, auxiliary structural controls are often incorporated into models. These controls help identify specific regions for editing, enabling precise local modifications while preserving the coherence of unedited regions with the original video. 
% This process, referred to as \textit{semantic decoupling}, is crucial for achieving high-performance video editing. 
However, existing methods often fail to achieve desirable structural controls.
Recent related studies can be categorized into two types:
\textbf{(1)} those relying on external control mechanisms, and 
\textbf{(2)} those relying on internal control mechanisms.
For the first type, methods resort to optical flow estimation~~\cite{yang2023rerender, liang2024flowvid}, depth estimation, edge estimation~~\cite{ouyang2024codef}, or semantic segmentation~~\cite{kahatapitiya2024object, esser2023structure}.
These methods have the following drawbacks:
\textbf{(1)} The above time-agnostic masks may not suit all editing tasks, such as shape editing, where the edited shape does not conform to the source shape mask.
\textbf{(2)} These methods require external pre-trained models, thereby introducing extra computational overhead for inference.
\textbf{(3)} Directly applying models designed for image processing to video processing often leads to significant artifacts~\cite{ouyang2024codef, yang2023rerender}, necessitating specialized designs and re-training~\cite{liang2024flowvid}.
These shortcomings make it a non-optimal choice for video editing.
The second type of video editing that relies on internal masks offers a more efficient solution.
The iconic method Prompt2prompt~~\cite{hertz2022prompt} has demonstrated the structure control capability of the cross-attention map.
Nevertheless, \textit{naively utilizing the imprecise masks derived from cross-attention maps leads to inferior video editing results}, which has been overlooked by previous research.

In response to this issue, we rethink the importance of attention masks in zero-shot video editing and propose \textbf{FreeMask} that leverages the attention masks by strategic mask usage.
FreeMask is based on \textit{two key observations} illustrated in Fig.~\ref{fig:insight1}:
\textbf{(1)} Cross-attention maps become clearer as denoising progresses. 
% as denoising steps increase
\textbf{(2)} Cross-attention maps exhibit the most precision in the middle layer, with the outer layer being too noisy and the inner layer being too imprecise due to low resolution.
To quantify these observations, we introduce Mask Matching Cost (MMC), an MIoU-based metric that measures layer-wise and timestep-wise attention mask precision, referred to as LMMC and TMMC, respectively.
% By these two metrics, we get quantitative observations and further design semantic-adaptive MMC for adaptive mask usage strategies tailored for different editing tasks, \textit{e.g.}, style translation, attribute editing, and shape editing, with different structural requirements adapting different optimal masks. 
Using these two metrics, we further design semantic-adaptive MMC for mask usage with varying precision control, tailored for different editing tasks, \textit{e.g.}, style translation, attribute editing, and shape editing.

% we identify the optimal layer and timestep for generating masks using cross-attention maps, given a specific video diffusion model. With the quantitative observations and metrics, We further design semantic-adaptive MMC for adaptive mask usage strategies tailored for different editing tasks, \textit{e.g.}, style translation, attribute editing, and shape editing.

% Finally, we identify a common issue in attention fusion during video editing --- \textit{over-constraint}, meaning that the source video overly restricts the editing target.
% We attribute this to imprecise mask guidance and partially masked fusion on a subset of attention types~\cite{qi2023fatezero}. We propose to integrate the strategic masks derived from MMC for guidance and perform feature blending in more comprehensive attention layer types --- temp-attention, cross-attention, and self-attention. This integration and comprehensive fusion substantially mitigate over-constraints and enhance the editing accuracy of FreeMask.

% Finally, we identify a common issue in attention fusion during zero-shot video editing --- \textit{over-constraint}. 
% This occurs when the source video overly restricts the editing process.
Moreover, most zero-shot video editing heavily relies on the blending of the source and edited attention features~\cite{ku2024anyv2v,qi2023fatezero,bai2024uniedit}. Nevertheless, determining the optimal blending ratio is challenging. Insufficient blending can lead to structural distortion, while excessive blending results in a completely identical video to the original, a common issue leading to prompt misalignment. 
To address this issue, we integrate MMC-selected masks with masked feature blending across various attention layer types—temporal, cross, and self-attention, as shown in Tab~\ref{table: compare}. This comprehensive masked fusion typically eliminates the need for ratio selection and significantly enhances the editing accuracy.
It is worth noting that FreeMask can be seamlessly integrated into existing zero-shot video editing frameworks without additional control assistance or parameter fine-tuning. The extensive experiments provide evidence of the numerous advantages FreeMask offers for zero-shot video editing without user-specific controls.

\section{Related Works}
\subsection{Video Editing}
%%%%%%0808version2%%%%%%%%%%%% conclusion driven
Recent advances in video generation have shifted from earlier efforts using deep generative models like GANs~\cite{goodfellow2020generative, tulyakov2018mocogan,saito2017temporal} to more recent diffusion models~\cite{ho2020denoising, Nichol_Dhariwal_2021, Song_Meng_Ermon_2020}, which produce higher-quality visual outputs. As diffusion models gain popularity, video editing techniques rapidly evolve to harness their potential. Current video editing methods can be broadly categorized into two types: \textbf{training-based methods} and \textbf{zero-shot methods}. As a consensus in many studies, video editing should satisfy the criteria of fidelity, alignment, and quality~\cite{xing2023survey}. 

\textbf{Training-based methods} enhance temporal consistency and motion fidelity by introducing temporal layers into pre-trained text-to-image (T2I) diffusion models~\cite{rombach2022high}. These methods train on large-scale datasets for generalizability or fine-tune specific video instances for efficiency. Models such as GEN-1~\cite{esser2023structure} and VideoComposer~\cite{wang2024videocomposer} exemplify the former, while Tune-A-Video~\cite{wu2023tune}, ControlVideo~\cite{zhang2023controlvideo}, and VideoP2P~\cite{liu2024video} represent the latter. Both approaches involve high training costs and may result in spatial detail distortion.

In contrast, \textbf{zero-shot methods} avoid extensive training by directly leveraging pre-trained diffusion models. Some early works based on pre-trained T2I models unavoidably introduce spatiotemporal distortion, such as Tokenflow~\cite{geyer2023tokenflow}, Rerender-A-Video~\cite{yang2023rerender}, CoDef~\cite{ouyang2024codef}, and FateZero~\cite{qi2023fatezero}. While some recent works significantly improved temporal consistency by utilizing open-source conditional video diffusion models like AnyV2V~\cite{ku2024anyv2v} and UniEdit~\cite{bai2024uniedit}, semantic fidelity remains challenging. Zero-shot methods often struggle to balance semantic fidelity and prompt alignment due to reliance on feature fusion between source and edited videos. To alleviate the restrictive fusion, mask guidance is widely employed.

\subsection{Mask Guidance for Video Editing}
%%%%%%%%%%%%%%%%%%0809%%%%%%%%version3%%%%%%%%%%%%
Mask guidance is an intuitive approach to facilitate strcutural control in video editing. Explicit segmentation masks enhance editing quality, as demonstrated by early works using generative adversarial networks~\cite{wang2018video, wang2019few} and advanced by recent diffusion-based methods like Text2Live~\cite{bar2022text2live}, Pix2Video~\cite{ceylan2023pix2video} and Object-Centric Diffusion~\cite{kahatapitiya2024object}. Despite their effectiveness, explicit masks can be resource-intensive and constrain shape modifications.

Alternatively, \textbf{cross-attention masks} offer a more flexible solution. Methods like VideoP2P~\cite{liu2024video} and FateZero~\cite{qi2023fatezero} use cross-attention masks inspired by Prompt2Prompt~\cite{hertz2022prompt} to improve spatiotemporal coherence while reducing reliance on external models. Although efficient, cross-attention masks may still produce artifacts if alignment variations are not addressed carefully.

\subsection{Other Structural Guidance for Video Editing}
%%%%%%%%%0809version2%%%%%%%%%

In addition to semantic masks, other structural information such as \textbf{depth/edge maps} can aid structural control. ControlNet~\cite{zhang2023adding} is widely used as a training-free plug-and-play module or integrated into video diffusion models for retraining. For training-free applications, examples include Rerender-A-Video~\cite{yang2023rerender}, ControlVideo~\cite{zhang2023controlvideo}, and MoonShot~\cite{zhang2024moonshot}. For retraining, examples are VideoComposer~\cite{wang2024videocomposer}, Gen-1~\cite{esser2023structure}, and Control-A-Video~\cite{chen2023control}. While training-free methods facilitate structural editing, they often struggle with temporal consistency. Retraining methods improve temporal coherence but are costly. Additionally, techniques like FLATTEN~\cite{cong2023flatten}, Rerender-A-Video~\cite{yang2023rerender}, CoDef~\cite{ouyang2024codef}, and FlowVid~\cite{liang2024flowvid} use \textbf{optical flow} for mask creation or canonical image construction. Overall, while these structural guidance methods improve video editing by preserving structure, they face challenges with shape modifications and require a balance between editing flexibility and structural accuracy.

\section{Preliminaries}
\label{sec: preliminaries}
\noindent\textbf{Text-to-video diffusion models.} Given a source video $\mathbf{X_0} \in \mathbb{R}^{F \times 3 \times H \times W}$ with $F$ frames of RGB images (height $H$ and width $W$ set to 512), a source prompt $P_0$, and an editing prompt $P_1$, we aim to generate an edited video $\mathbf{X_1}$ using text-to-video diffusion models. We encode $\mathbf{X_0}$ into a latent $\mathbf{Z_0} \in \mathbb{R}^{F \times d \times h \times w}$ using an image encoder $\mathcal{E}$, where $d$ is the latent dimension and $h, w$ are 64. The latent $\mathbf{Z_0}$ can be decoded back to $\mathbf{\widehat{X}_0} \approx \mathbf{X_0}$ using a decoder $\mathcal{D}$.

During inference, a 3D-Unet~\cite{cciccek20163d} $\epsilon_{\theta}$ denoises the latent variable $\mathbf{Z}_t$ using the noise schedule $\bar{\alpha}_t$ and a text embedding $\psi(P)$ from a text encoder $\psi$. For DDIM sampling~\cite{Song_Meng_Ermon_2020}, the latent variable $\mathbf{Z}_{t-1}$ is updated at timestep t from $\mathbf{Z}_t$ by: $\mathbf{Z}_{t-1} = \sqrt{\bar{\alpha}_{t-1}} \left( \frac{\mathbf{Z}_t - \sqrt{1 - \bar{\alpha}_t} \cdot \epsilon_{\theta}(\mathbf{Z}_t, \psi(P), t)}{\sqrt{\bar{\alpha}_t}} \right) + \sqrt{1 - \bar{\alpha}_{t-1} - \sigma_t^2} \cdot \epsilon_{\theta}(\mathbf{Z}_t, \psi(P), t)$, where $\sigma_t$ denotes the added noise, usually set to 0 for deterministic sampling. Repeating this process for $T$ steps yields the final clean latent.

\noindent\textbf{DDIM inversion. }The reverse operation of DDIM sampling~\cite{Song_Meng_Ermon_2020}, known as DDIM inversion~\cite{mokady2023null}, estimates $\mathbf{Z}_{t+1}$ from $\mathbf{Z}_t$: $\mathbf{Z}_{t+1} = \sqrt{\frac{\bar{\alpha}_{t+1}}{\bar{\alpha}_t}} \mathbf{Z}_t + \left(\sqrt{\frac{1}{\bar{\alpha}_{t+1}} - 1} - \sqrt{\frac{1}{\bar{\alpha}_t} - 1}\right) \cdot \epsilon_{\theta}(\mathbf{Z}_t, \psi(P), t)$. Repeating this process for $T$ steps yields the final noisy latent variable $\mathbf{Z}_T$, which is then used as the initial latent for the denoising process. 
% After DDIM inversion in video diffusion models, the initial latent retains more motion information than in image diffusion models, which can be used to alleviate the issue of temporal inconsistency.
 
% \subsection{Cross-attention Masks in Diffusion Models} 
\noindent\textbf{Cross-attention masks in diffusion models. }To ensure the generated video aligns with the text description, the model leverages cross-attention~\cite{Wei_Zhang_Li_Zhang_Wu_2020} during denoising. Specifically, the deep spatial-temporal features of the noisy video frame $\mathbf{\tilde{Z}_t}$, are projected into a query matrix $Q_C = \ell_{Q_C}(\mathbf{\tilde{Z}_t}) \in \mathbb{R}^{F \times S \times d'}$, where $S$ is sequence length of $Q_C$ and $d'$ is the embedding dimension. The text embedding is projected into a key matrix $K_C = \ell_{K_C}(\psi(P)) \in \mathbb{R}^{F \times S' \times d'}$  and a value matrix $V_C = \ell_{V_C}(\psi(P)) \in \mathbb{R}^{F \times S' \times d'} $ using learned linear projections $\ell_{Q_C}, \ell_{K_C}, \ell_{V_C}$, where $S'$ is sequence length of $K_C$ and $V_C$. The \textit{cross-attention maps} are given by:

\begin{equation}
A_C = \text{Softmax}\left( \frac{{Q_C}{K_C}^T}{ \sqrt{d'}}\right)
\end{equation}
% where $d'$ denotes the embedding dimension of the keys and queries, $S$ represents the sequence length of the text embeddings, and $S'$ corresponds to the sequence length of the video frame feature patches. 
% where $d'$ denotes the embedding dimension of the keys and queries. 
As a whole, $A_C \in \mathbb{R}^{F \times S \times S'} $ reflects the similarity between $Q_C$ and $K_C$, which enables correlations between text embedding and semantic layout. The binary masks are obtained by thresholding $A_C$ with a constant $\tau$~\cite{avrahami2022blended,qi2023fatezero}.

% Theoretically, the accuracy of text-video alignment in cross-attention depends on factors including model weights, text embeddings, and image features.

\section{Methodology}
\subsection{Method Overview}
\begin{figure*}[tbp] 
  \centering
  \includegraphics[width=\textwidth]{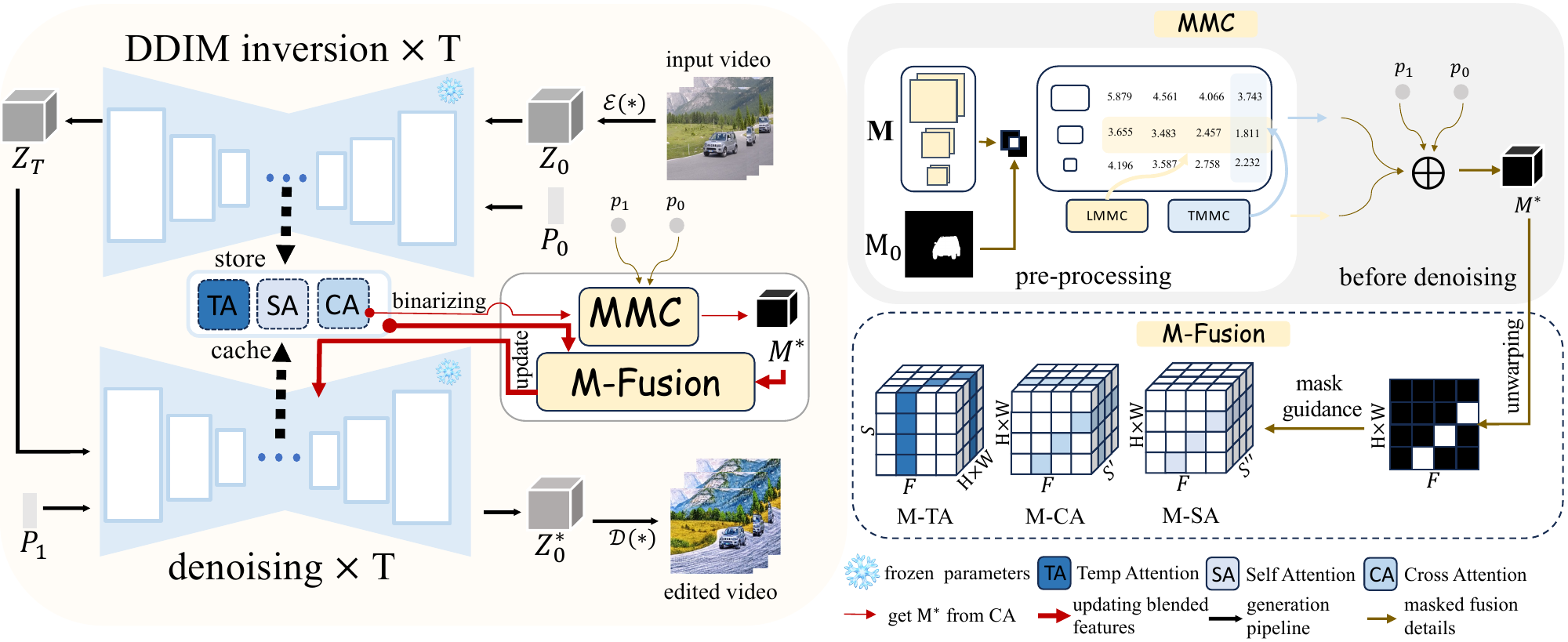} % 替换为您的图片文件名
  \caption{FreeMask overview. FreeMask takes source video $\mathbf{X_0}$  and text prompt $P_0$ as input. During preprocessing, it stores cross-attention maps for each timestep across all videos in the DAVIS testing dataset to calculate LMMC and TMMC. In the inference stage, $\mathbf{X_0}$ and $P_0$ are input to DDIM inversion, storing attention features at each timestep and collecting the final latent output as the initial latent for denoising. Before denoising, masks $\mathbf{M^*}$ are adaptively promoted. During denoising, attention features are blended using masks $\mathbf{M^*}$. The final latent output $\mathbf{Z_0^*}$ is then decoded to produce the edited video.}
  % \hangjie{1. I did not find the meaning of $F_S$, $F_T$ in the body text. 2. It is hard to grasp what Mask-guided Attention Fusion does after observing the illustration.}
  % }\modified{modified later}
  % \vspace{-.3cm}
  \label{fig: framework}
\end{figure*}

% \textbf{FreeMask} aims at high-performance prompt-driven zero-shot video editing without user-defined controls. To this end, we redesign the selection and usage strategies of attention masks to enhance \textit{semantic decoupling} and alleviate \textit{over-constraint}.
As illustrated in Fig.~\ref{fig: framework}, we leverage the well-established inversion-then-denoising pipelines~\cite{avrahami2022blended,qi2023fatezero,ku2024anyv2v} and cached attention features belonging to 3D-Unet of all inversion steps for editing. 
In the pre-processing step, we first calculate the model-dependent LMMC and TMMC metrics using the DAVIS video dataset. During video editing, after performing DDIM inversion and before denoising, we compute the semantic-adaptive MMC metrics to select attention masks, as detailed inSec~\ref{sec: MMC}. During denoising, we apply our masks to guide the fusion of different types of attention features, as detailed in Sec~\ref{sec: mask_guidance}.

\subsection{ Mask Matching Cost (MMC)}
\label{sec: MMC}
\noindent\textbf{Key observations.} As mentioned above and shown in Fig.~\ref{fig:insight1} (a), we observe that the semantic clarity of cross-attention maps is not constant but varies with model structure and denoising timesteps. Furthermore, this structural-aware and timestep-aware semantic clarity is more pronounced in video diffusion models than in image diffusion models. 
To quantify these observations, we define the metric Mask Matching Cost (MMC) to identify and select masks with precision control.

\noindent\textbf{Mask candidates extraction.} 
Before quantitative observations, 
we first generate mask candidates $\mathbf{M} = \left\{ M_t^l \right\}_{t=0,l=0}^{T-1, L-1}$,  where $L$ represents the number of cross-attention layers in the diffusion U-Net, $T$ denotes the number of sampling steps, and ${M}_t^l \in \mathbb{B}^{F \times H \times W}$ represents the mask candidate at the $l$-th layer and timestep $t$.
% we first generate mask candidates $\mathbf{M}\in \mathbb{B}^{T \times L \times F \times H \times W}$, where $L$ represents the number of cross-attention layers in the diffusion U-Net, $T$ denotes the number of sampling steps, and ${M}_t^l \in \mathbb{B}^{F \times H \times W}$ represents the mask candidate at the $l$-th layer and timestep $t$. 
These candidates are derived from cross-attention maps obtained during the DDIM inversion steps, as illustrated in Fig.~\ref{fig: framework}. Specifically, for each cross-attention map, we extract the submap corresponding to the object word $p_0$ in the prompt $P_0$.
% denoted as $\mathbf{CA_0} = \left\{ ca_t^i \right\}_{t=0,i=0}^{T-1, L-1}$, where $ca_t^i$ represents the cross-attention map at timestep $t$ and layer $i$. 
These maps are then binarized using a threshold hyperparameter $\tau$, as described in \cite{qi2023fatezero,avrahami2022blended}, and reshaped to match the resolution of the input video frames, resulting in the final mask candidates $\mathbf{M}$.

\noindent\textbf{Layer-wise and Timestep-wise Mask Matching Cost.} 
%%%%%%%%%%%version 2 %%%%%%%%%% 0807 
To quantify the structural and temporal diversity of cross-attention maps, we introduce the Layer-wise Mask Matching Cost (LMMC) and Timestep-wise Mask Matching Cost (TMMC). We use a video dataset with dense annotations and per-frame ground truth segmentation (DAVIS dataset~\cite{Perazzi_CVPR_2016} for practice). For a given video $\mathbf{X_0}$ from the dataset, we use the corresponding ground-truth segmentation masks as reference masks, denoted as $M_0 \in \mathbb{B}^{F \times H \times W}$. We then compute the Mean Intersection over Union (MIoU) between $\mathbf{M}$ and $M_0$ to evaluate the matching performance: 
\begin{equation}
\label{eq:matching}
    d_t^l = \frac{M_0 \cap M_t^l}{|M_t^l| + |M_0| - |M_0 \cap M_t^l|},
\end{equation}
where $d_t^l$ denotes the MIoU of $M_t^l$ and $M_0$. Then we define:
\begin{equation}
\label{eq:matching_s}
\mathcal{D}_{LMMC} = \left\{ D_l = \frac{1}{T} \sum_{t=1}^{T} \frac{1}{d_t^l} \,\middle|\, l = 0, 1, \ldots, L-1 \right\}
\end{equation}

\begin{equation}
\label{eq:matching_s_min}
l^* = \mathrm{argmin}_{D_l \in \mathcal{D}_{LMMC}} D_l
\end{equation}
where $\mathcal{D}_{LMMC}$ denotes the set of LMMC across layers, and $l^*$ represents the layer index with minimum LMMC.

%%%%%%%%%%%%version1 %%%%%%%%%%%%Why do we gather, calculate and define it?
% We gather pre-semantic masks from a pre-processed dataset and concatenate masks from all frames as the video mask reference $M_0 = [m^0, \ldots, m^F]$. Subsequently, masks in $\textbf{M}$ are interpolated to $\textbf{M}'$ across different layers and resolutions, aligning them with the spatial resolution of $M_0$. Then we calculate the Mean Intersection over Union (MIoU):
% \begin{equation}
% \label{eq:matching}
%     d_t^l = \frac{M_0 \cap M_i^l}{|M_t^l| + |M_0| - |M_0 \cap M_t^l|},
% \end{equation}
% where $d_t^l$ denote the MIoU metric values at timestep $t$ in the $l$-th cross-attention layer. 

% Using $d_t^l$, we define LMMC as:
% \begin{equation}
% \label{eq:matching_s}
%     \hat{\sigma}_s =\mathrm{argmin}_{\sigma_s \in \{0, 1, \ldots, L\}} \frac{1}{T} \sum_{t=1}^{T} \frac{1}{d_t^l},
% \end{equation}
% where $\hat{\sigma}_s$ denotes the optimal cross-attention index selected to minimize the average costs across all $T$ timesteps.

% \noindent\textbf{Timestep-wise MMC.} 
Similar to the LMMC, we define the TMMC as:
\begin{equation}
\label{eq:matching_t}
\mathcal{D}_{TMMC} = \left\{ D_t=\frac{1}{L} \sum_{l=1}^{L} \frac{1}{d_t^l} \,\middle|\, t = 0, 1, \ldots, T-1 \right\}
\end{equation}

\begin{equation}
\label{eq:matching_t_min}
t^* = \mathrm{argmin}_{D_t \in \mathcal{D}_{TMMC}} D_t
\end{equation}
where $\mathcal{D}_{TMMC}$ is the set of TMMC across timsteps, and $t^*$ is timestep with minimum TMMC.

\noindent\textbf{Rethinking cross-attention matching for video and text. }
%目的：更佳细化和强调观察并且给出理论依据结论，为下文做铺垫
Using the designed metrics, we quantify our observations and provide a theoretical explanation. We analyzed four T2V models (Lavie~\cite{wang2023lavie}, modelscope~\cite{wang2023modelscopet2v}, zeroscope~\cite{zeroscopev2}, and a pseudo-T2V model of Fatezero~\cite{qi2023fatezero}) with 60 prompt-video pairs and 20 randomly selected video samples with per-frame single instance segmentation masks from DAVIS~\cite{Perazzi_CVPR_2016}.
% We found that the cross-attention mask variety is common in video diffusion models and systematic with the model framework, meaning it relates to the model architecture rather than the input videos.
We found that the clarity variety of cross-attention masks is common in video diffusion models and systematic at the model level, meaning the regularity of the variety is related to model architecture rather than input videos. 
The details of the key observations are illustrated in Fig.~\ref{fig:insight1} (b) and (c): 
(1)\textit{ Cross-attention matching accuracy exhibits an inverted U-shape with increasing layers.} 
% The alignment between text and image is less precise in the outer layers compared to the inner layers. 
% This discrepancy is due to outer layers containing more low-level spatial information, while inner layers capture higher-level semantic information. Effective video-text matching requires alignment between text embeddings and latent information at similar levels. 
This discrepancy stems from that the outer layers contain low-level spatial information while the inner layers capture higher-level spatial information. Text embeddings are high-level information, and the similarity of semantic level is a prerequisite for effective video-text matching, as the cross-attention map calculates the matric product of the query from text embeddings and the key from video latent.
Nonetheless, semantic matching accuracy does not equate to mask matching accuracy, as the resolution in the innermost layer is too low, leading to higher LMMC.
(2)\textit{ Matching accuracy increases with denoising timesteps.} Timesteps in the denoising process run in reverse order, with T decreasing towards 0. Initially, high latent noise results in less accurate low-level spatial information and, consequently, less precise high-level semantic matching.

\noindent\textbf{Semantic-adaptive Mask Matching Cost. }
After quantifying the structure-aware and timestep-aware semantic disparity with LMMC and TMMC separately, we select the optimal masks to guide the editing process. Let $\mathbf{M^*} = \left\{ {\widehat{M}}_t^l \right\}_{t=0,i=0}^{T-1, L-1}$ be the optimal masks, where $\widehat{M}_t^l \in \mathbb{B}^{F \times H \times W}$ represents the optimal mask at the $l$-th layer and timestep $t$. We highlight this systematic semantic disparity in cross-attention as an advantage of mask precision control to meet the needs of different tasks. Time-agnostic accurate masks are essential for tasks demanding high structural coherence. Conversely, time-aware (coarse-to-fine) masks are needed for tasks like shape editing that involve structural transformation. While previous work relies on accurate segmentation masks and coarse bounding boxes for mask precision control~\cite{xie2023smartbrush}, the intrinsic properties of cross-attention can achieve this more efficiently and effectively.
To this end, we leverage TMMC and LMMC to design the semantic-adaptive MMC,  which enables adaptively choosing time-agnostic and time-aware masks through a designed Kronecker delta function $\delta$ :

\begin{equation}
\delta = 
\begin{cases} 
1 & \text{if } p_0 = p_1 \\
0 & \text{if } p_0 \neq p_1
\end{cases}
\end{equation}

\begin{equation}
\widehat{M}_t^l = 
\begin{cases} 
M_t^{l^*},  \quad \text{if } \delta =0, \\
M_{t^*}^{l^*}, \quad \text{if } \delta=1.
\end{cases}
\end{equation}
 where $p_1$ corresponds to the object in editing prompt $P_1$. For tasks like shape editing, $p_1 \neq p_0$ and time-aware masks are used for tasks requiring flexible shape changes. Otherwise, time-agnostic masks are employed for tasks demanding high structural coherence. 
 % With these selective masks, we effectively apply mask guidance to different attention mechanisms.

%%%%%%%%%%version1%%%%%%%%% % Why do we integrate? Is it necessary?
% Finally, we integrate TMMC ($\hat{\sigma}_t$) and LMMC ($\hat{\sigma}_s$) to get MMC $\hat{\sigma}$:
% \begin{equation}
% \label{eq:mmc}
%    \{\hat{r},\hat{\sigma}\} = \lambda\{\hat{r}_t,{\hat\sigma}_t\} + \{\hat{r}_s, \hat{\sigma}_s\}
% \end{equation}
% where $\lambda = p_1 \oplus p_0$ and $p_1$ corresponds to the object in editing prompt $P_1$. When $\lambda = 1$, it signifies using time-aware masks for tasks requiring flexible shape change. Conversely, time-agnostic masks are employed for tasks requiring high structural coherence. To achieve high editing quality, we note that in the MMC selection stage, mask precision in video editing should vary depending on the task type: precise masks are crucial for tasks requiring structural coherence, such as stylization and attribute editing, while shape editing benefits from a mix of fine and coarse masks.

\subsection{Applications of Mask Guidance}
\label {sec: mask_guidance}
% Unlike previous work with partially masked fusion on subset types of attention modules, as shown in Tab.~\ref{table: compare}, we 
% integrate MMC-selected masks with masked feature blending across comprehensive attention layer types—temporal, cross, and self-attention. This approach alleviates the need for ratio selection while enhancing the editing quality.

Unlike previous studies that use partially masked fusion for a subset of attention types, as shown in Tab.~\ref{table: compare}, we apply MMC-selected masks to all major attention types to alleviate blending over-constraint while enhancing editing quality. 
 
\noindent\textbf{Temp-attention feature blending. } Temp-attention is crucial for maintaining temporal motion consistency~~\cite{bai2024uniedit, ku2024anyv2v}. It tracks pixel-level motion by capturing relative transformations between pixels in different frames. However, an excessive fusion of this fine-grained motion information leads to an edited video identical to the source video. To address the over-constraint, we use masks to decouple the edited region within the temp-attention features, where the batch size matches the number of input feature tokens. First, at timestep $t$ of the denoising process, we convert the masks $\mathbf{M_t^*}\in \mathbb{B}^{F \times H \times W}$ into:
\begin{equation}
\mathbf{m^*_t} = \mathcal{F}((1-\delta)\mathbf{M^*_t})
\label{eq:3-1}
\end{equation}
where the mask $\mathbf{M^*_t}$ is reshaped on spatial dimension and flattened into a two-dimensional tensor $\mathbf{m^*} \in \mathbb{B}^{h'w' \times F} $ using the unwrapping function $\mathcal{F}$, where $h'\times w'$ equals to the batch size of temp-attention, which is also the resolution of the image features. For tasks preserving the entire structure like stylization, $\delta = 1$ and $\mathbf{m^*_t}=0$. We apply $\mathbf{m^*_t}$ to the bath size dimension of temp-attention:
\begin{equation}
  \{K_{T}^{*}, Q_{T}^{*}\} =  (1-\mathbf{m^*})  \odot \{K_{T}^{0}, Q_{T}^{0}\} + \mathbf{m^*}\odot \{K_{T}^{1}, Q_{T}^{1}\}
  \label{eq:3-2}
\end{equation}
where \(K_{T}^{0}, Q_{T}^{0}, K_{T}^{1}, Q_{T}^{1}, K_{T}^{*}, Q_{T}^{*}\) represent the source temp-key and temp-query, the edited temp-key and temp-query, and the blended temp-key and temp-query, respectively, all of which belong to the space \(\mathbb{R}^{h'w' \times F \times \text{dim}}\), and \(\odot\) denotes the Hadamard product. 
% This fusion process selectively blends features based on the mask, allowing precise attention modifications while decoupling the edited object's motion information. 
% However, it does not imply random motion in the masked area. Instead, cross-attention and initial latent representations maintain some motion information at a coarser level, preserving the overall movement rather than pixel-level details.

\noindent\textbf{Cross-attention feature blending.} To preserve structural details, cross-attention maps are often reweighted, refined, or replaced with source maps~\cite{hertz2022prompt, liu2024video, qi2023fatezero}. Despite the efficacy, these methods overlook the influence of neighboring function words, which can retain some spatial information about the source object and reduce editing accuracy, e.g., in the prompt "a jeep driving in the countryside," the cross-attention for "a" or "driving" may still show the "jeep" contour, impacting accuracy. To mitigate this issue, we introduce orthogonal masked blending to enhance standard blending operations. We unwarp $\mathbf{M^*_t}$ into $\mathbf{m^*_t}$ with Eq.~\ref{eq:3-1} and blend it with the source cross-attention before performing usual fusion operations:
\begin{equation}
  \mathbf{A_C^{*}} =  (1-\mathbf{m^*})  \odot \mathbf{A_C^0} + \mathbf{m^*}\odot \mathbf{A_C^1}
  \label{eq:3-3}
\end{equation}
where $\mathbf{A_C^{*}, A_C^0, A_C^1}$ are the blended, source and edited cross-attentions respectively, belonging to $\mathbb{R}^F \times h'w' \times S'$.

\noindent\textbf{Self-attention feature blending. }We follow the masked self-attention feature blending approach in \cite{hertz2022prompt, qi2023fatezero,ku2024anyv2v}, but extend masked self-attention blending from original attribute and shape editing to stylization. For stylization, masks are applied inversely, with $\mathbf{m^*}$ masking source self-attention maps. We observe that stylization may lead to undesirable deformation in moving objects, which can be alleviated by masked self-attention blending.

% \hangjie{Details of the strategies for different editing tasks? }
% \reply{in semantic-adaptive mask matching cost.}

\begin{table}[t]

  \centering
  \resizebox{\linewidth}{!}{
  \begin{tabular}{cllllllll}
  
  \toprule
  Methods    & SM &TM &M-SA & M-CA  & M-TA & M-L & zero-shot &backbone \\

  \midrule
   Tokenflow  &\textcolor{red}{\ding{55}} &\textcolor{red}{\ding{55}} &\textcolor{red}{\ding{55}} &\textcolor{red}{\ding{55}}  &\textcolor{red}{\ding{55}} &\textcolor{red}{\ding{55}} &\textcolor{darkgreen}{\ding{51}}  & Stable Diffusion   \\
   Pix2Video  &\textcolor{red}{\ding{55}} &\textcolor{red}{\ding{55}} &\textcolor{red}{\ding{55}} &\textcolor{red}{\ding{55}}  &\textcolor{red}{\ding{55}} &\textcolor{red}{\ding{55}} &\textcolor{darkgreen}{\ding{51}}  & SD-depth   \\
   Rerender-A-Video  &\textcolor{red}{\ding{55}}  &\textcolor{red}{\ding{55}} &\textcolor{red}{\ding{55}} &\textcolor{red}{\ding{55}}  &\textcolor{red}{\ding{55}} &\textcolor{darkgreen}{\ding{51}} &\textcolor{darkgreen}{\ding{51}}   & Controlnet   \\
   Text2Video-Zero   &\textcolor{red}{\ding{55}} &\textcolor{red}{\ding{55}} &\textcolor{red}{\ding{55}} &\textcolor{red}{\ding{55}}  &\textcolor{red}{\ding{55}} &\textcolor{darkgreen}{\ding{51}} &\textcolor{darkgreen}{\ding{51}} & Stable Diffusion   \\
   FateZero   &\textcolor{red}{\ding{55}} &\textcolor{darkgreen}{\ding{51}} &\textcolor{darkgreen}{\ding{51}} &\textcolor{red}{\ding{55}}  &\textcolor{red}{\ding{55}} &\textcolor{red}{\ding{55}} &w/o shape & Stable Diffusion   \\
   CoDeF &\textcolor{red}{\ding{55}}  &\textcolor{red}{\ding{55}} &\textcolor{red}{\ding{55}} &\textcolor{red}{\ding{55}}  &\textcolor{red}{\ding{55}} &\textcolor{red}{\ding{55}}  &\textcolor{red}{\ding{55}}   & ControlNet  \\
   Tune-A-Video &\textcolor{red}{\ding{55}}  &\textcolor{red}{\ding{55}} &\textcolor{red}{\ding{55}} &\textcolor{red}{\ding{55}}  &\textcolor{red}{\ding{55}} &\textcolor{red}{\ding{55}} &\textcolor{red}{\ding{55}}   & Stable Diffusion   \\

   Video-P2P  &\textcolor{red}{\ding{55}} &\textcolor{darkgreen}{\ding{51}} &\textcolor{red}{\ding{55}} &\textcolor{red}{\ding{55}}  &\textcolor{red}{\ding{55}} &\textcolor{darkgreen}{\ding{51}} &\textcolor{red}{\ding{55}}  & Stable Diffusion   \\
  \midrule

  Ours     &\textcolor{darkgreen}{\ding{51}}  &\textcolor{darkgreen}{\ding{51}} &\textcolor{darkgreen}{\ding{51}} &\textcolor{darkgreen}{\ding{51}}  &\textcolor{darkgreen}{\ding{51}} &\textcolor{darkgreen}{\ding{51}} &\textcolor{darkgreen}{\ding{51}}   & Any T2V   \\
  
  \bottomrule
  \end{tabular}}
  \caption{
  % \textbf{Comparision with different video editing methods.} \hangjie{Explain the meaning of the abbreviation M, SA, CA and TA.}}
  Comparision with different video editing methods. 
  M, SA, CA, TA, and L abbreviate masks, self-attention, cross-attention, temporal attention, and latent, respectively. 
  SM and TM refer to masks that account for structural and timestep-wise variations, respectively.}
  \label{table: compare}
  \end{table}
% problem: reference citations are too long.

\section{Experiments}

\subsection{Experimental Setup}
% \noindent\textbf{Datasets.} We evaluated our method on public DAVIS~\cite{Perazzi_CVPR_2016} videos and Internet videos depicting various moving subjects. To assess performance, we tested on 8-frame videos with a resolution of 512 × 512, using a variety of text prompts to produce different editing results for each video. Our quantitative evaluation dataset comprises 60 text-video pairs.
% \hangjie{Please specify the number of Internet videos used and their specific sources. Moreover, I suggest moving the introduction of metrics used into this part and naming this part \textbf{Datasets and evaluation metrics}.}

\noindent\textbf{Datasets and evaluation metrics. }We evaluated our method on public DAVIS~\cite{Perazzi_CVPR_2016} videos and 20 Internet videos from pexels~\cite{pexels} depicting various moving subjects.  To assess performance, we tested 8-frame videos with a resolution of 512 × 512, using a variety of text prompts to produce different editing results for each video. 
% Our quantitative evaluation dataset comprises 60 text-video pairs. For quantitative comparison, we conducted human evaluations to assess editing quality ('Edit'), image fidelity ('Image'), and temporal consistency ('Temp'). Additionally, we computed CLIP~\cite{Radford_Kim_Hallacy_Ramesh_Goh_Agarwal_Sastry_Amanda_Mishkin_Clark_et} scores for temporal consistency ('CLIP-Img') and editing accuracy ('CLIP-Text'). For 'CLIP-Text', we calculated the average cosine similarity between CLIP image embeddings of all frames and the CLIP text embedding of the edit prompt. For 'CLIP-Img', we measured the average cosine similarity between image embeddings of consecutive frames. 
We report four metrics for quantitative comparison and three metrics for user study. For quantitative comparison, we compute CLIP~\cite{Radford_Kim_Hallacy_Ramesh_Goh_Agarwal_Sastry_Amanda_Mishkin_Clark_et} scores for temporal consistency ('Temp') and video-text alignment ('CLIP'), following~\cite{qi2023fatezero}. Additionally, we calculate Masked PSNR ('M.PSNR') and LPIPS~\cite{zhang2018unreasonable} for structure preservation, following~\cite{liu2024video}. For more detailed explanations of the metrics and annotation process, please refer to the supplementary materials.

% For 'Temp', we calculated the average cosine similarity between CLIP image embeddings of all frames and the CLIP text embedding of the edit prompt. For 'Clip', we measured the average cosine similarity between image embeddings of consecutive frames. 
% For user study, we assess video alignment to editing prompt('Edit'), image fidelity ('Image'), and temporal consistency ('Quality'). 

\noindent\textbf{Implementation details.}
\begin{figure*}[tbp] % 使用 figure* 环境横跨两栏
  \centering % 图片居中
  \includegraphics[width=\textwidth]{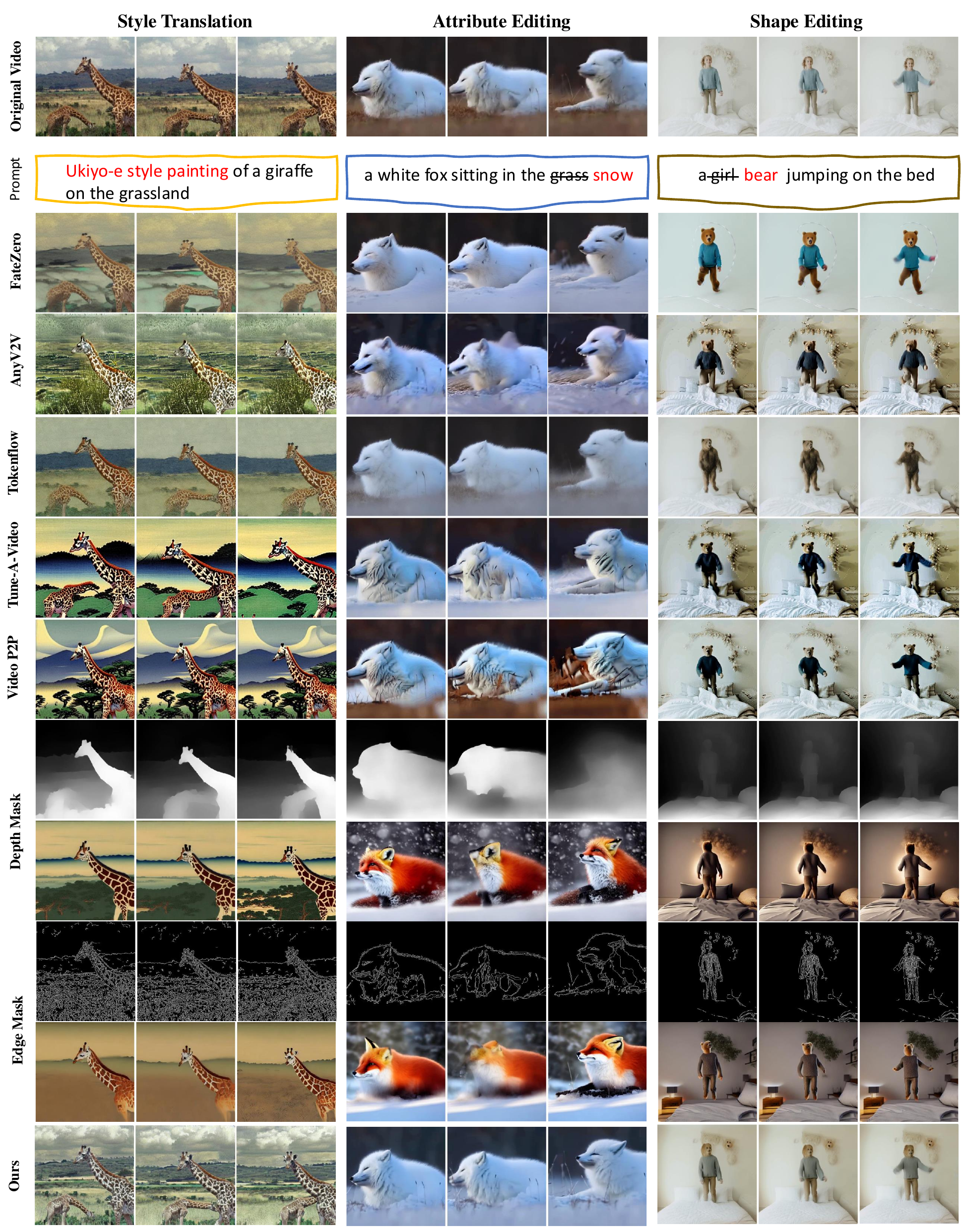} % 替换为您的图片文件名
  \caption{Comparison results with several state-of-the-art approaches on three distinct tasks: stylization, attribute editing, and shape editing.} % 图片下方的说明
  \label{fig:compare1} % 为图片定义标签，便于引用
\end{figure*}
We evaluated our zero-shot video editing approach using the pre-trained T2V model Zeroscope~\cite{zeroscopev2}. For each source video, we perform 50 steps of DDIM inversion~\cite{Song_Meng_Ermon_2020}, followed by generating outputs using DDIM deterministic sampling~\cite{Song_Meng_Ermon_2020} with classifier-free guidance~\cite{ho2022classifier} at a scale of 7.5. 
This process takes 1 minute on an NVIDIA A100 GPU. 
% \hangjie{Please be precise on the time taken for this process. You can clearly get a precise number for this, right?}
% During sampling, self-attention, temporal attention, cross-attention, and the final latent output of each step are fused differently based on the specific editing task, such as style translation, attribute editing, or shape editing.
During sampling, attention features are fused with selected masks. The fusion strategies and masks are tailored to specific tasks like stylization, attribute editing and shape editing.
% \hangjie{I think the first part of this sentence is rather redundant.}
\begin{figure*}[tbp] % 使用 figure* 环境横跨两栏
  \centering % 图片居中
  \includegraphics[width=\textwidth]{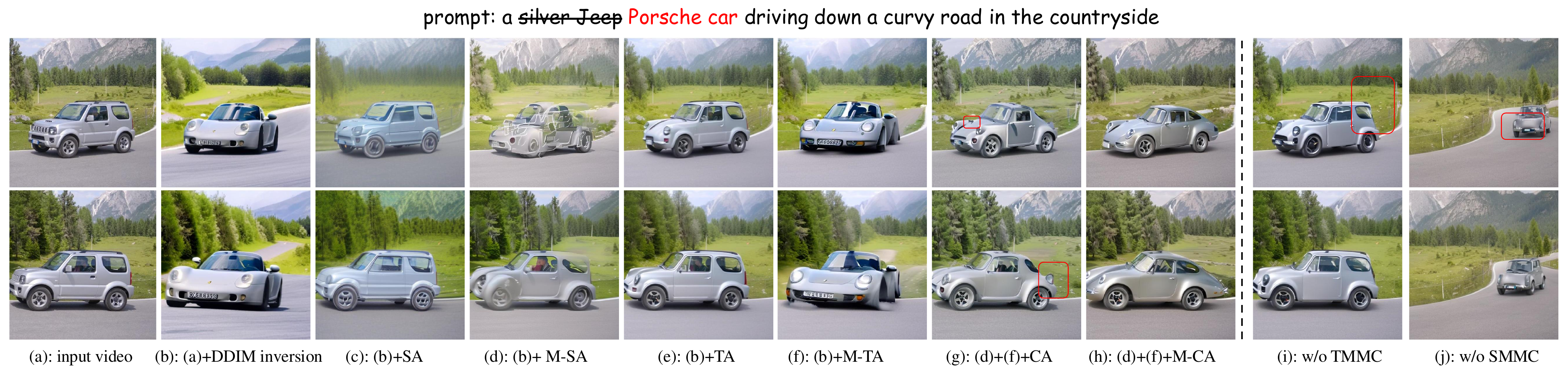} 
  \caption{Ablation experiments. The experiments use Zeroscope~\cite{zeroscopev2} as the base model and are conducted on a shape-editing task that changes a Jeep into a Porsche car. In these experiments. (a) is the original input video; (b) shows results using the latent output from DDIM inversion as the initial latent; (c) represents the fusion of self-attention based on (b); (d) involves the fusion of masked self-attention based on (b). Similar logic applies to other sub-captions.}
  \label{fig:exp_ablation2} % 为图片定义标签，便于引用
  \vspace{-.3cm}
\end{figure*}
  % \hangjie{The captions are so confusing that I can barely understand the ablation study (like two figure (a)?).
  % What's worse, your references in Sec. 5.3 is confusing as well.
  % Btw, you did not explain what is the baseline.}
\subsection{Comparison of Video Editing}

\noindent\textbf{Qualitative comparison results.} 
 We compare FreeMask with state-of-the-art methods, including FateZero~\cite{qi2023fatezero}, TokenFlow~\cite{geyer2023tokenflow}, ControlVideo~\cite{zhang2023controlvideo} using optimized depth and edge masks, and AnyV2V~\cite{ku2024anyv2v} leveraging I2V model. Additionally, we compared it with fine-tuned methods like Tune-A-Video~\cite{wu2023tune} and Video-P2P~\cite{liu2024video}.

Figure~\ref{fig:compare1} presents the comparison results. For the \textbf{stylization} task (first column), our method aligns texture and color with the prompt \textit{"Ukiyo-e style painting"} while maintaining high structural and temporal coherence. Other methods struggled with content retention and detail preservation, particularly for moving objects in the middle of the video.
In the \textbf{attribute editing} task (second column), our method excels by transforming grass into snowy grass according to the prompt \textit{"snow"} while preserving the fine details of the grass blades.
For the \textbf{shape editing} task (third column), our zero-shot method successfully changes the shape to match the prompt \textit{"bear"} while retaining the original structure, motion, and facial expressions. Other zero-shot methods had difficulty with object motion and consistency. Although fine-tuned methods performed well, they required fine-tuning and still struggled with preserving motion details.

 % \modified{In \textbf{stylization}, Pix2Video~\cite{ceylan2023pix2video}, Tune-A-Video~\cite{wu2023tune}, VideoP2P~\cite{liu2024video}, and ControlVideo~\cite{zhang2023controlvideo} struggled with content retention, while TokenFlow~\cite{geyer2023tokenflow} and FateZero~\cite{qi2023fatezero} showed some improvement but lacked consistency and detailed artistic texture. Our method excelled in content, structural, and temporal consistency, producing clear images with a well-transferred artistic style. }
 
 % For \textbf{attribute editing}, our approach effectively transformed grass into snowy grass without losing detail. For \textbf{shape editing}, FateZero~\cite{qi2023fatezero} and TokenFlow~\cite{geyer2023tokenflow} struggled with object motion and consistency, while Tune-A-Video~\cite{wu2023tune} and VideoP2P~\cite{liu2024video} excelled but required fine-tuning. Our zero-shot model demonstrated superior shape editing with greater efficiency than fine-tuned methods.

\noindent\textbf{Quantitative comparison and user study results.} 
As shown in Tab~\ref{table: quantitative_baseline}, our method significantly outperforms others in temporal consistency, video-text alignment and structure preservation
in quantitative comparison, and demonstrates superior performance in editing accuracy, image quality, and temporal consistency in user studies. Please refer to the supplementary materials for details.

% \hangjie{You need to provide more details on the user study, including the number of annotators involved, the annotation process, the score scale, the meaning of three aspects, what kind of videos can be annotated with high scores in those aspects, etc. Currently, they are all missing from your manuscript. I would suggest adding them and moving part of them to your Appendix. Some reviewers are quite picky about user study since CLIP metrics indeed can not reveal the quality of the videos. You can even introduce more video-based metrics.}

\begin{table}[t]

  \centering
  \resizebox{\linewidth}{!}{
  \begin{tabular}{@{}l@{\hspace{2mm}}c@{\hspace{2mm}}*{3}{c@{\hspace{2mm}}}c@{\hspace{2mm}}c@{\hspace{2mm}}c@{\hspace{2mm}}}
  
  \toprule

  Method & \multicolumn{4}{c}{Quantitative Comparison}& \multicolumn{3}{c}{User Study} \\
  \cmidrule(l{1mm}r{1mm}){2-5} 
  \cmidrule(l{1mm}r{1mm}){6-8} 
   & Temp$\textcolor{red}{\uparrow}$ & CLIP$\textcolor{red}{\uparrow}$ & M.PSNR$\textcolor{red}{\uparrow}$ & LPIPS$\textcolor{darkgreen}{\downarrow}$ &Edit$\textcolor{red}{\uparrow}$ & Image$\textcolor{red}{\uparrow}$ & Quality$\textcolor{red}{\uparrow}$\\

  \midrule
  \small{FateZero}                 & 0.937 & 0.278         &22.72  &0.383  &6.12 &8.96 &7.12 \\
  \small{Tokenflow}                & 0.941 & 0.273         &25.41  &0.381 &6.88 &9.01 &7.56 \\
  \small{Video-P2P}                & 0.923 & 0.284         &18.71  &0.508 &5.22 &5.92 &7.35\\
  \small{Tune-A-Video}             & 0.928 & \textbf{0.285}         &17.31  &0.514  &5.14 &5.88 &7.43\\
  \small{ControlVideo}             & 0.916 & 0.281         &18.55  &0.539 &7.99 &6.67 &7.12 \\
  \small{AnyV2V}                   & 0.921 & 0.271         &18.23  &0.497 &5.96 &6.74 &7.08\\
  \small{FateZero + MMC}             & 0.939 & 0.279         &23.25  &0.381  &6.55 &8.97 &7.18 \\
  \midrule 
  Ours                             & \textbf{0.952} & 0.282 &\textbf{25.94}  &\textbf{0.366}  &\textbf{7.68} &\textbf{9.23} &\textbf{8.90} \\
  
  \bottomrule
  \end{tabular}
  }
  \caption{Quantitative comparison and user study results.}
  \label{table: quantitative_baseline}
  \end{table}
%question: if citations are added, it seems too long.
  % \hangjie{\textbf{Quantitative evaluation and user study of FreeMask against baselines.}} 
  % In our user study, the results of our method are preferred over those from baselines. For CLIP-Score, we achieve the best temporal consistency and comparable framewise editing accuracy against an optimization-based image editing method.
  % \hangjie{It seems that the explanation of the results should be moved to the body text. Moreover, I do not understand the meaning of "Inversion and editing".}
  % }

\begin{figure}[t] 
  \centering 
  \includegraphics[width=\linewidth]{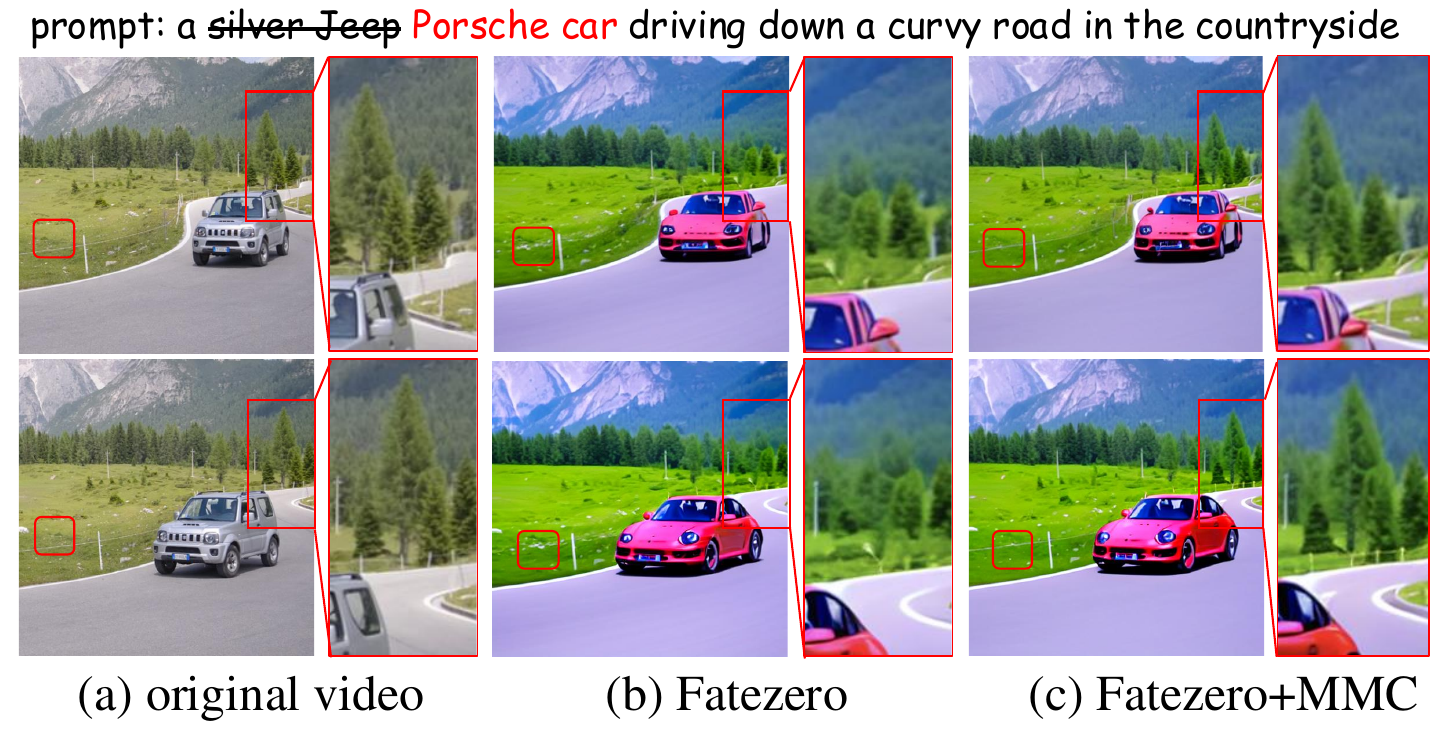}
  \caption{Extension results on shape editing.} 
  \vspace{-.3cm}
  \label{fig: extension1} 
\end{figure}

\subsection{Ablation Study}
\noindent\textbf{Effect of individual cost of MMC. } 
As shown in Fig.~\ref{fig:exp_ablation2}, (i) demonstrates mask-guided fusion without TMMC, leading to rigid transitions from 'Jeep' to 'Porsche' and unnatural changes, particularly in the tailstock. (j) shows mask-guided fusion without LMMC, resulting in noticeable distortion of structural details, especially in small objects.

\noindent\textbf{Effect of MMC on different fusion mechanisms.}  We further investigate the design of mask guidance in the fusion of different features. As indicated in Fig.~\ref{fig:exp_ablation2}, blending masked attentions (including temporal, self, and cross attentions) effectively decouples unedited attributes (dynamic information in temporal attention, structure information in self-attention, and boundary information in cross-attention).

\subsection{Extending FreeMask to Other Methods}
\noindent\textbf{Qualitative Results.} We tested our mask selection using MMC metrics on FateZero~\cite{qi2023fatezero}, a zero-shot video editing model based on a flattened text-to-image framework with masked self-attention for shape editing. As shown in Fig.~\ref{fig: extension1}, MMC-selected masks enhance structural details, especially around moving objects, while the original masks cause flickering due to suboptimal masking. 

\noindent\textbf{Quantitative results.}
As shown in Tab.~\ref{table: quantitative_baseline}, we compare FateZero~\cite{qi2023fatezero} before and after adding MMC-selected masks quantitatively and can find it helps with temporal consistency and image editing quality. Additional comparisons, ablation, extension studies and full videos are available in the supplementary materials.

% \section{Conclusion and Future Work}
\section{Conclusion}
% In this paper, we propose $\textbf{FreeMask}$. We observe and leverage the systematic semantic disparity in cross-attention to enhance masked feature fusion for zero-shot video editing. This innovative approach enhances editing precision and alleviates attribute over-constraints, improving semantic clarity and reducing artifacts.

% In this paper, we identify the issue of semantic disparity in zero-shot video editing by the proposed MMC metric and propose $\textbf{FreeMask}$ that selects quality masks for more comprehensive feature blending.
% $\textbf{FreeMask}$ alleviates semantic disparity and feature over-constraints, improving semantic clarity and reducing artifacts for various video editing tasks.

Despite the efficacy of cross-attention mask controls in image editing,  naively utilizing them leads to inferior video editing results. We identify that these issues stem from semantic timestep-wise and model layer-wise variations in cross-attention maps. We leverage the systematic varieties and propose $\textbf{FreeMask}$ that selects optimal masks for more comprehensive feature blending, alleviating semantic disparity and blending over-constraints, improving semantic clarity, and reducing artifacts for various video editing tasks.

\appendix
\bibliography{aaai25}\label{sec:reference_examples}

\begin{thebibliography}{49}
\providecommand{\natexlab}[1]{#1}

\bibitem[{Avrahami, Lischinski, and Fried(2022)}]{avrahami2022blended}
Avrahami, O.; Lischinski, D.; and Fried, O. 2022.
\newblock Blended diffusion for text-driven editing of natural images.
\newblock In \emph{Proceedings of the IEEE/CVF conference on computer vision and pattern recognition}, 18208--18218.

\bibitem[{Bai et~al.(2024)Bai, He, Wang, Guo, Hu, Liu, and Bian}]{bai2024uniedit}
Bai, J.; He, T.; Wang, Y.; Guo, J.; Hu, H.; Liu, Z.; and Bian, J. 2024.
\newblock Uniedit: A unified tuning-free framework for video motion and appearance editing.
\newblock \emph{arXiv preprint arXiv:2402.13185}.

\bibitem[{Bar-Tal et~al.(2022)Bar-Tal, Ofri-Amar, Fridman, Kasten, and Dekel}]{bar2022text2live}
Bar-Tal, O.; Ofri-Amar, D.; Fridman, R.; Kasten, Y.; and Dekel, T. 2022.
\newblock Text2live: Text-driven layered image and video editing.
\newblock In \emph{European conference on computer vision}, 707--723. Springer.

\bibitem[{Cerspense(2023)}]{zeroscopev2}
Cerspense. 2023.
\newblock ZeroScope v2 - 576w.

\bibitem[{Ceylan, Huang, and Mitra(2023)}]{ceylan2023pix2video}
Ceylan, D.; Huang, C.-H.~P.; and Mitra, N.~J. 2023.
\newblock Pix2video: Video editing using image diffusion.
\newblock In \emph{Proceedings of the IEEE/CVF International Conference on Computer Vision}, 23206--23217.

\bibitem[{Chen et~al.(2023)Chen, Ji, Wu, Wu, Xie, Li, Xia, Xiao, and Lin}]{chen2023control}
Chen, W.; Ji, Y.; Wu, J.; Wu, H.; Xie, P.; Li, J.; Xia, X.; Xiao, X.; and Lin, L. 2023.
\newblock Control-a-video: Controllable text-to-video generation with diffusion models.
\newblock \emph{arXiv preprint arXiv:2305.13840}.

\bibitem[{{\c{C}}i{\c{c}}ek et~al.(2016){\c{C}}i{\c{c}}ek, Abdulkadir, Lienkamp, Brox, and Ronneberger}]{cciccek20163d}
{\c{C}}i{\c{c}}ek, {\"O}.; Abdulkadir, A.; Lienkamp, S.~S.; Brox, T.; and Ronneberger, O. 2016.
\newblock 3D U-Net: learning dense volumetric segmentation from sparse annotation.
\newblock In \emph{Medical Image Computing and Computer-Assisted Intervention--MICCAI 2016: 19th International Conference, Athens, Greece, October 17-21, 2016, Proceedings, Part II 19}, 424--432. Springer.

\bibitem[{Cong et~al.(2023)Cong, Xu, Simon, Chen, Ren, Xie, Perez-Rua, Rosenhahn, Xiang, and He}]{cong2023flatten}
Cong, Y.; Xu, M.; Simon, C.; Chen, S.; Ren, J.; Xie, Y.; Perez-Rua, J.-M.; Rosenhahn, B.; Xiang, T.; and He, S. 2023.
\newblock Flatten: optical flow-guided attention for consistent text-to-video editing.
\newblock \emph{arXiv preprint arXiv:2310.05922}.

\bibitem[{Esser et~al.(2023)Esser, Chiu, Atighehchian, Granskog, and Germanidis}]{esser2023structure}
Esser, P.; Chiu, J.; Atighehchian, P.; Granskog, J.; and Germanidis, A. 2023.
\newblock Structure and content-guided video synthesis with diffusion models.
\newblock In \emph{Proceedings of the IEEE/CVF International Conference on Computer Vision}, 7346--7356.

\bibitem[{Federico~Perazzi et~al.(2016)Federico~Perazzi, McWilliams, Gool, Gross, and Sorkine-Hornung}]{Perazzi_CVPR_2016}
Federico~Perazzi, J. P.-T.; McWilliams, B.; Gool, L.~V.; Gross, M.; and Sorkine-Hornung, A. 2016.
\newblock A Benchmark Dataset and Evaluation Methodology for Video Object Segmentation.
\newblock In \emph{The IEEE Conference on Computer Vision and Pattern Recognition (CVPR)}.

\bibitem[{Geyer et~al.(2023)Geyer, Bar-Tal, Bagon, and Dekel}]{geyer2023tokenflow}
Geyer, M.; Bar-Tal, O.; Bagon, S.; and Dekel, T. 2023.
\newblock Tokenflow: Consistent diffusion features for consistent video editing.
\newblock \emph{arXiv preprint arXiv:2307.10373}.

\bibitem[{Goodfellow et~al.(2020)Goodfellow, Pouget-Abadie, Mirza, Xu, Warde-Farley, Ozair, Courville, and Bengio}]{goodfellow2020generative}
Goodfellow, I.; Pouget-Abadie, J.; Mirza, M.; Xu, B.; Warde-Farley, D.; Ozair, S.; Courville, A.; and Bengio, Y. 2020.
\newblock Generative adversarial networks.
\newblock \emph{Communications of the ACM}, 63(11): 139--144.

\bibitem[{Hertz et~al.(2022)Hertz, Mokady, Tenenbaum, Aberman, Pritch, and Cohen-Or}]{hertz2022prompt}
Hertz, A.; Mokady, R.; Tenenbaum, J.; Aberman, K.; Pritch, Y.; and Cohen-Or, D. 2022.
\newblock Prompt-to-prompt image editing with cross attention control.
\newblock \emph{arXiv preprint arXiv:2208.01626}.

\bibitem[{Ho et~al.(2022)Ho, Chan, Saharia, Whang, Gao, Gritsenko, Kingma, Poole, Norouzi, Fleet et~al.}]{ho2022imagen}
Ho, J.; Chan, W.; Saharia, C.; Whang, J.; Gao, R.; Gritsenko, A.; Kingma, D.~P.; Poole, B.; Norouzi, M.; Fleet, D.~J.; et~al. 2022.
\newblock Imagen video: High definition video generation with diffusion models.
\newblock \emph{arXiv preprint arXiv:2210.02303}.

\bibitem[{Ho, Jain, and Abbeel(2020)}]{ho2020denoising}
Ho, J.; Jain, A.; and Abbeel, P. 2020.
\newblock Denoising diffusion probabilistic models.
\newblock \emph{Advances in neural information processing systems}, 33: 6840--6851.

\bibitem[{Ho and Salimans(2022)}]{ho2022classifier}
Ho, J.; and Salimans, T. 2022.
\newblock Classifier-free diffusion guidance.
\newblock \emph{arXiv preprint arXiv:2207.12598}.

\bibitem[{Kahatapitiya et~al.(2024)Kahatapitiya, Karjauv, Abati, Porikli, Asano, and Habibian}]{kahatapitiya2024object}
Kahatapitiya, K.; Karjauv, A.; Abati, D.; Porikli, F.; Asano, Y.~M.; and Habibian, A. 2024.
\newblock Object-centric diffusion for efficient video editing.
\newblock \emph{arXiv preprint arXiv:2401.05735}.

\bibitem[{Khachatryan et~al.(2023)Khachatryan, Movsisyan, Tadevosyan, Henschel, Wang, Navasardyan, and Shi}]{khachatryan2023text2video}
Khachatryan, L.; Movsisyan, A.; Tadevosyan, V.; Henschel, R.; Wang, Z.; Navasardyan, S.; and Shi, H. 2023.
\newblock Text2video-zero: Text-to-image diffusion models are zero-shot video generators.
\newblock In \emph{Proceedings of the IEEE/CVF International Conference on Computer Vision}, 15954--15964.

\bibitem[{Ku et~al.(2024)Ku, Wei, Ren, Yang, and Chen}]{ku2024anyv2v}
Ku, M.; Wei, C.; Ren, W.; Yang, H.; and Chen, W. 2024.
\newblock Anyv2v: A plug-and-play framework for any video-to-video editing tasks.
\newblock \emph{arXiv preprint arXiv:2403.14468}.

\bibitem[{Liang et~al.(2024)Liang, Wu, Wang, Yu, Li, Zhao, Misra, Huang, Zhang, Vajda et~al.}]{liang2024flowvid}
Liang, F.; Wu, B.; Wang, J.; Yu, L.; Li, K.; Zhao, Y.; Misra, I.; Huang, J.-B.; Zhang, P.; Vajda, P.; et~al. 2024.
\newblock Flowvid: Taming imperfect optical flows for consistent video-to-video synthesis.
\newblock In \emph{Proceedings of the IEEE/CVF Conference on Computer Vision and Pattern Recognition}, 8207--8216.

\bibitem[{Liu et~al.(2024)Liu, Zhang, Li, Lin, and Jia}]{liu2024video}
Liu, S.; Zhang, Y.; Li, W.; Lin, Z.; and Jia, J. 2024.
\newblock Video-p2p: Video editing with cross-attention control.
\newblock In \emph{Proceedings of the IEEE/CVF Conference on Computer Vision and Pattern Recognition}, 8599--8608.

\bibitem[{Mokady et~al.(2023)Mokady, Hertz, Aberman, Pritch, and Cohen-Or}]{mokady2023null}
Mokady, R.; Hertz, A.; Aberman, K.; Pritch, Y.; and Cohen-Or, D. 2023.
\newblock Null-text inversion for editing real images using guided diffusion models.
\newblock In \emph{Proceedings of the IEEE/CVF Conference on Computer Vision and Pattern Recognition}, 6038--6047.

\bibitem[{Nichol and Dhariwal(2021)}]{Nichol_Dhariwal_2021}
Nichol, A.; and Dhariwal, P. 2021.
\newblock Improved Denoising Diffusion Probabilistic Models.
\newblock \emph{Cornell University - arXiv,Cornell University - arXiv}.

\bibitem[{Ouyang et~al.(2024)Ouyang, Wang, Xiao, Bai, Zhang, Zheng, Zhou, Chen, and Shen}]{ouyang2024codef}
Ouyang, H.; Wang, Q.; Xiao, Y.; Bai, Q.; Zhang, J.; Zheng, K.; Zhou, X.; Chen, Q.; and Shen, Y. 2024.
\newblock Codef: Content deformation fields for temporally consistent video processing.
\newblock In \emph{Proceedings of the IEEE/CVF Conference on Computer Vision and Pattern Recognition}, 8089--8099.

\bibitem[{Pexels(2024)}]{pexels}
Pexels. 2024.
\newblock {Pexels Stock Photos}.
\newblock \url{https://www.pexels.com}.
\newblock [Accessed: 2024-08-05].

\bibitem[{Pont-Tuset et~al.(2017)Pont-Tuset, Perazzi, Caelles, Arbeláez, Sorkine-Hornung, and Gool}]{DAVIS2017}
Pont-Tuset, J.; Perazzi, F.; Caelles, S.; Arbeláez, P.; Sorkine-Hornung, A.; and Gool, L. 2017.
\newblock The 2017 DAVIS Challenge on Video Object Segmentation.
\newblock \emph{Cornell University - arXiv,Cornell University - arXiv}.

\bibitem[{Qi et~al.(2023)Qi, Cun, Zhang, Lei, Wang, Shan, and Chen}]{qi2023fatezero}
Qi, C.; Cun, X.; Zhang, Y.; Lei, C.; Wang, X.; Shan, Y.; and Chen, Q. 2023.
\newblock Fatezero: Fusing attentions for zero-shot text-based video editing.
\newblock In \emph{Proceedings of the IEEE/CVF International Conference on Computer Vision}, 15932--15942.

\bibitem[{Radford et~al.(2021)Radford, Kim, Hallacy, Ramesh, Goh, Agarwal, Sastry, Amanda, Mishkin, Clark, Krueger, and Sutskever}]{Radford_Kim_Hallacy_Ramesh_Goh_Agarwal_Sastry_Amanda_Mishkin_Clark_et}
Radford, A.; Kim, J.; Hallacy, C.; Ramesh, A.; Goh, G.; Agarwal, S.; Sastry, G.; Amanda, A.; Mishkin, P.; Clark, J.; Krueger, G.; and Sutskever, I. 2021.
\newblock Learning Transferable Visual Models From Natural Language Supervision.
\newblock \emph{Cornell University - arXiv,Cornell University - arXiv}.

\bibitem[{Rombach et~al.(2022)Rombach, Blattmann, Lorenz, Esser, and Ommer}]{rombach2022high}
Rombach, R.; Blattmann, A.; Lorenz, D.; Esser, P.; and Ommer, B. 2022.
\newblock High-resolution image synthesis with latent diffusion models.
\newblock In \emph{Proceedings of the IEEE/CVF conference on computer vision and pattern recognition}, 10684--10695.

\bibitem[{Saito, Matsumoto, and Saito(2017)}]{saito2017temporal}
Saito, M.; Matsumoto, E.; and Saito, S. 2017.
\newblock Temporal generative adversarial nets with singular value clipping.
\newblock In \emph{Proceedings of the IEEE international conference on computer vision}, 2830--2839.

\bibitem[{Simonyan and Zisserman(2014)}]{simonyan2014very}
Simonyan, K.; and Zisserman, A. 2014.
\newblock Very deep convolutional networks for large-scale image recognition.
\newblock \emph{arXiv preprint arXiv:1409.1556}.

\bibitem[{Song, Meng, and Ermon(2020)}]{Song_Meng_Ermon_2020}
Song, J.; Meng, C.; and Ermon, S. 2020.
\newblock Denoising Diffusion Implicit Models.
\newblock \emph{arXiv: Learning,arXiv: Learning}.

\bibitem[{Tulyakov et~al.(2018)Tulyakov, Liu, Yang, and Kautz}]{tulyakov2018mocogan}
Tulyakov, S.; Liu, M.-Y.; Yang, X.; and Kautz, J. 2018.
\newblock Mocogan: Decomposing motion and content for video generation.
\newblock In \emph{Proceedings of the IEEE conference on computer vision and pattern recognition}, 1526--1535.

\bibitem[{Wang et~al.(2023{\natexlab{a}})Wang, Yuan, Chen, Zhang, Wang, and Zhang}]{wang2023modelscopet2v}
Wang, J.; Yuan, H.; Chen, D.; Zhang, Y.; Wang, X.; and Zhang, S. 2023{\natexlab{a}}.
\newblock Modelscope text-to-video technical report.
\newblock \emph{arXiv preprint arXiv:2308.06571}.

\bibitem[{Wang et~al.(2019)Wang, Liu, Tao, Liu, Kautz, and Catanzaro}]{wang2019few}
Wang, T.-C.; Liu, M.-Y.; Tao, A.; Liu, G.; Kautz, J.; and Catanzaro, B. 2019.
\newblock Few-shot video-to-video synthesis.
\newblock \emph{arXiv preprint arXiv:1910.12713}.

\bibitem[{Wang et~al.(2018)Wang, Liu, Zhu, Liu, Tao, Kautz, and Catanzaro}]{wang2018video}
Wang, T.-C.; Liu, M.-Y.; Zhu, J.-Y.; Liu, G.; Tao, A.; Kautz, J.; and Catanzaro, B. 2018.
\newblock Video-to-video synthesis.
\newblock \emph{arXiv preprint arXiv:1808.06601}.

\bibitem[{Wang et~al.(2024)Wang, Yuan, Zhang, Chen, Wang, Zhang, Shen, Zhao, and Zhou}]{wang2024videocomposer}
Wang, X.; Yuan, H.; Zhang, S.; Chen, D.; Wang, J.; Zhang, Y.; Shen, Y.; Zhao, D.; and Zhou, J. 2024.
\newblock Videocomposer: Compositional video synthesis with motion controllability.
\newblock \emph{Advances in Neural Information Processing Systems}, 36.

\bibitem[{Wang et~al.(2023{\natexlab{b}})Wang, Chen, Ma, Zhou, Huang, Wang, Yang, He, Yu, Yang et~al.}]{wang2023lavie}
Wang, Y.; Chen, X.; Ma, X.; Zhou, S.; Huang, Z.; Wang, Y.; Yang, C.; He, Y.; Yu, J.; Yang, P.; et~al. 2023{\natexlab{b}}.
\newblock Lavie: High-quality video generation with cascaded latent diffusion models.
\newblock \emph{arXiv preprint arXiv:2309.15103}.

\bibitem[{Wei et~al.(2020)Wei, Zhang, Li, Zhang, and Wu}]{Wei_Zhang_Li_Zhang_Wu_2020}
Wei, X.; Zhang, T.; Li, Y.; Zhang, Y.; and Wu, F. 2020.
\newblock Multi-Modality Cross Attention Network for Image and Sentence Matching.
\newblock In \emph{2020 IEEE/CVF Conference on Computer Vision and Pattern Recognition (CVPR)}.

\bibitem[{Wu et~al.(2023)Wu, Ge, Wang, Lei, Gu, Shi, Hsu, Shan, Qie, and Shou}]{wu2023tune}
Wu, J.~Z.; Ge, Y.; Wang, X.; Lei, S.~W.; Gu, Y.; Shi, Y.; Hsu, W.; Shan, Y.; Qie, X.; and Shou, M.~Z. 2023.
\newblock Tune-a-video: One-shot tuning of image diffusion models for text-to-video generation.
\newblock In \emph{Proceedings of the IEEE/CVF International Conference on Computer Vision}, 7623--7633.

\bibitem[{Xie et~al.(2023)Xie, Zhang, Lin, Hinz, and Zhang}]{xie2023smartbrush}
Xie, S.; Zhang, Z.; Lin, Z.; Hinz, T.; and Zhang, K. 2023.
\newblock Smartbrush: Text and shape guided object inpainting with diffusion model.
\newblock In \emph{Proceedings of the IEEE/CVF Conference on Computer Vision and Pattern Recognition}, 22428--22437.

\bibitem[{Xing et~al.(2023)Xing, Feng, Chen, Dai, Hu, Xu, Wu, and Jiang}]{xing2023survey}
Xing, Z.; Feng, Q.; Chen, H.; Dai, Q.; Hu, H.; Xu, H.; Wu, Z.; and Jiang, Y.-G. 2023.
\newblock A survey on video diffusion models.
\newblock \emph{arXiv preprint arXiv:2310.10647}.

\bibitem[{Yang et~al.(2023)Yang, Zhou, Liu, and Loy}]{yang2023rerender}
Yang, S.; Zhou, Y.; Liu, Z.; and Loy, C.~C. 2023.
\newblock Rerender a video: Zero-shot text-guided video-to-video translation.
\newblock In \emph{SIGGRAPH Asia 2023 Conference Papers}, 1--11.

\bibitem[{Yuan et~al.(2024)Yuan, Zhang, Wang, Wei, Feng, Pan, Zhang, Liu, Albanie, and Ni}]{yuan2024instructvideo}
Yuan, H.; Zhang, S.; Wang, X.; Wei, Y.; Feng, T.; Pan, Y.; Zhang, Y.; Liu, Z.; Albanie, S.; and Ni, D. 2024.
\newblock InstructVideo: instructing video diffusion models with human feedback.
\newblock In \emph{Proceedings of the IEEE/CVF Conference on Computer Vision and Pattern Recognition}, 6463--6474.

\bibitem[{Zhang et~al.(2024)Zhang, Li, Le, Shou, Xiong, and Sahoo}]{zhang2024moonshot}
Zhang, D.~J.; Li, D.; Le, H.; Shou, M.~Z.; Xiong, C.; and Sahoo, D. 2024.
\newblock Moonshot: Towards controllable video generation and editing with multimodal conditions.
\newblock \emph{arXiv preprint arXiv:2401.01827}.

\bibitem[{Zhang, Rao, and Agrawala(2023)}]{zhang2023adding}
Zhang, L.; Rao, A.; and Agrawala, M. 2023.
\newblock Adding conditional control to text-to-image diffusion models.
\newblock In \emph{Proceedings of the IEEE/CVF International Conference on Computer Vision}, 3836--3847.

\bibitem[{Zhang et~al.(2018)Zhang, Isola, Efros, Shechtman, and Wang}]{zhang2018unreasonable}
Zhang, R.; Isola, P.; Efros, A.~A.; Shechtman, E.; and Wang, O. 2018.
\newblock The unreasonable effectiveness of deep features as a perceptual metric.
\newblock In \emph{Proceedings of the IEEE conference on computer vision and pattern recognition}, 586--595.

\bibitem[{Zhang et~al.(2023{\natexlab{a}})Zhang, Wang, Zhang, Zhao, Yuan, Qin, Wang, Zhao, and Zhou}]{zhang2023i2vgen}
Zhang, S.; Wang, J.; Zhang, Y.; Zhao, K.; Yuan, H.; Qin, Z.; Wang, X.; Zhao, D.; and Zhou, J. 2023{\natexlab{a}}.
\newblock I2vgen-xl: High-quality image-to-video synthesis via cascaded diffusion models.
\newblock \emph{arXiv preprint arXiv:2311.04145}.

\bibitem[{Zhang et~al.(2023{\natexlab{b}})Zhang, Wei, Jiang, Zhang, Zuo, and Tian}]{zhang2023controlvideo}
Zhang, Y.; Wei, Y.; Jiang, D.; Zhang, X.; Zuo, W.; and Tian, Q. 2023{\natexlab{b}}.
\newblock Controlvideo: Training-free controllable text-to-video generation.
\newblock \emph{arXiv preprint arXiv:2305.13077}.

\end{thebibliography}

\clearpage
\section*{Appendix}
% \hangjie{In this appendix, we ﬁrst discuss the XXX (Appendix A) and XXX (Appendix B), then ...}
In this appendix, we first discuss additional experiment details (Appendix~\ref{appendix1}), including details of quantitative and human evaluations, additional ablation, comparison, and extension results. We then discuss limitations and potential future work.
\section{Additional Experiment Details}
\label{appendix1}
\subsection{Quantitative  and Human Evaluations}
We evaluate FreeMask with 4 quantitative metrics, including Temporal Consistency ('Temp'), Clip Score ('Clip'), Masked PSNR ('M.PSNR'), and Perceptual Similarity ('LPIPS') and 3 metrics for user study, including Text Alignment ('Edit'), Image Fidelity ('Img') and Temporal Consistency and Realism ('Quality'). 

\noindent\textbf{CLIP scores. }
We choose the official ViT-Large-Patch 14 CLIP model and use the output logits as the CLIP Score output. For Clip Score, we calculate the average cosine similarity between CLIP~\cite{Radford_Kim_Hallacy_Ramesh_Goh_Agarwal_Sastry_Amanda_Mishkin_Clark_et} image embeddings of all frames and the CLIP text embedding of the edit prompt. For Temporal Consistency, we calculate the average cosine similarity between CLIP~\cite{Radford_Kim_Hallacy_Ramesh_Goh_Agarwal_Sastry_Amanda_Mishkin_Clark_et} image embeddings of consecutive frames of edited videos.

\noindent\textbf{Masked PSNR \&LPIPS. }
To evaluate the structural preservation, we calculate the Masked PSNR following~\cite{liu2024video}, but the annotation process is a little different with VideoP2P~\cite{liu2024video}. Masked PSNR measures the low-level pixel distance in unedited regions. In our experiments, for tasks like stylization that includes global texture and color change but preserves all structure details, we regard all-region for Masked PSNR; for tasks like attribute and shape editing that include local editing, we regard the unedited region for Masked PSNR. For this calculation, we use 10 videos from DAVIS~\cite{DAVIS2017} and employ 30 prompts for stylization, attribute editing, and shape editing. Then, we use the pre-semantic segmentation annotations M from DAVIS~\cite{DAVIS2017} and calculate $M^*=\neg M$ to precisely locate the unedited region and calculate the PSNR of the edited video $\mathbf{X_1}$ and the source video $\mathbf{X_0}$:

\begin{equation}
    M.PSNR(\mathbf{X_1},\mathbf{X_0}) =PSNR(B(\mathbf{X_1},M^*), B(\mathbf{X_1},M^*)
    \label{eq:3-3}
\end{equation}
where $B(\dots)$ is the binary operation with a threshold of 0.3. 
For LIPIS, we follow ~\cite{zhang2018unreasonable} and apply the standard VGG~\cite{simonyan2014very} for feature extraction.

\noindent\textbf{User Study. }
We conducted a user study by distributing a form to users to visually compare our method with other methods. For Text Alignment, we asked users if the edited videos were aligned with the editing prompt, especially the edited region. For Image Fidelity, we asked users if the unedited regions of edited videos were coherent with the original video. For Quality, we asked users the temporal consistency and realism quality (especially less flickering and smoother edge of edited and edited regions). We collect the results of the quantitative study, including 30 pairwise and another 6 internet videos from pexels~\cite{pexels} with 5 prompts separately, totaling 60 pairwise comparisons. And we distribute it to 20 Internet users, totaling 1200 comparison responses. The score scale ranges from 0 to 10, with 10 being the highest, satisfying the criteria. For each method, we exclude the highest and lowest scores for each video-text pair and calculate the average of the remaining scores to obtain the final score for that method on this metric.

\subsection{More Experiment Setting Details. }
The experiments were conducted using PyTorch version 2.1, CUDA version 1.18 and Python 3.11. Computational resources included an NVIDIA A100 GPU with 80GB memory. For hyperparameters, our proposed design includes four key hyperparameters: $\alpha_s \in [1, T], \alpha_c \in [1, T],  \alpha_t \in [1, T] $ represent the final timestep for masked blending of self-attention, cross-attention and temp-attention separately. Lower $\alpha$ values indicate greater integration of inversion attention. As mentioned in the paper, we avoid tedious hyperparameter tuning through comprehensive masked fusion, setting $\alpha_s, \alpha_c,  \alpha_t $ all to 0.99. In addition, we adopt hyperparameter $\tau$ that thresholds attention masks as 0.3 following ~\cite{qi2023fatezero}.

\subsection{Additional Comparison Results}
\begin{figure*}[tbp] % 使用 figure* 环境横跨两栏
  \centering % 图片居中
  \includegraphics[width=\textwidth]{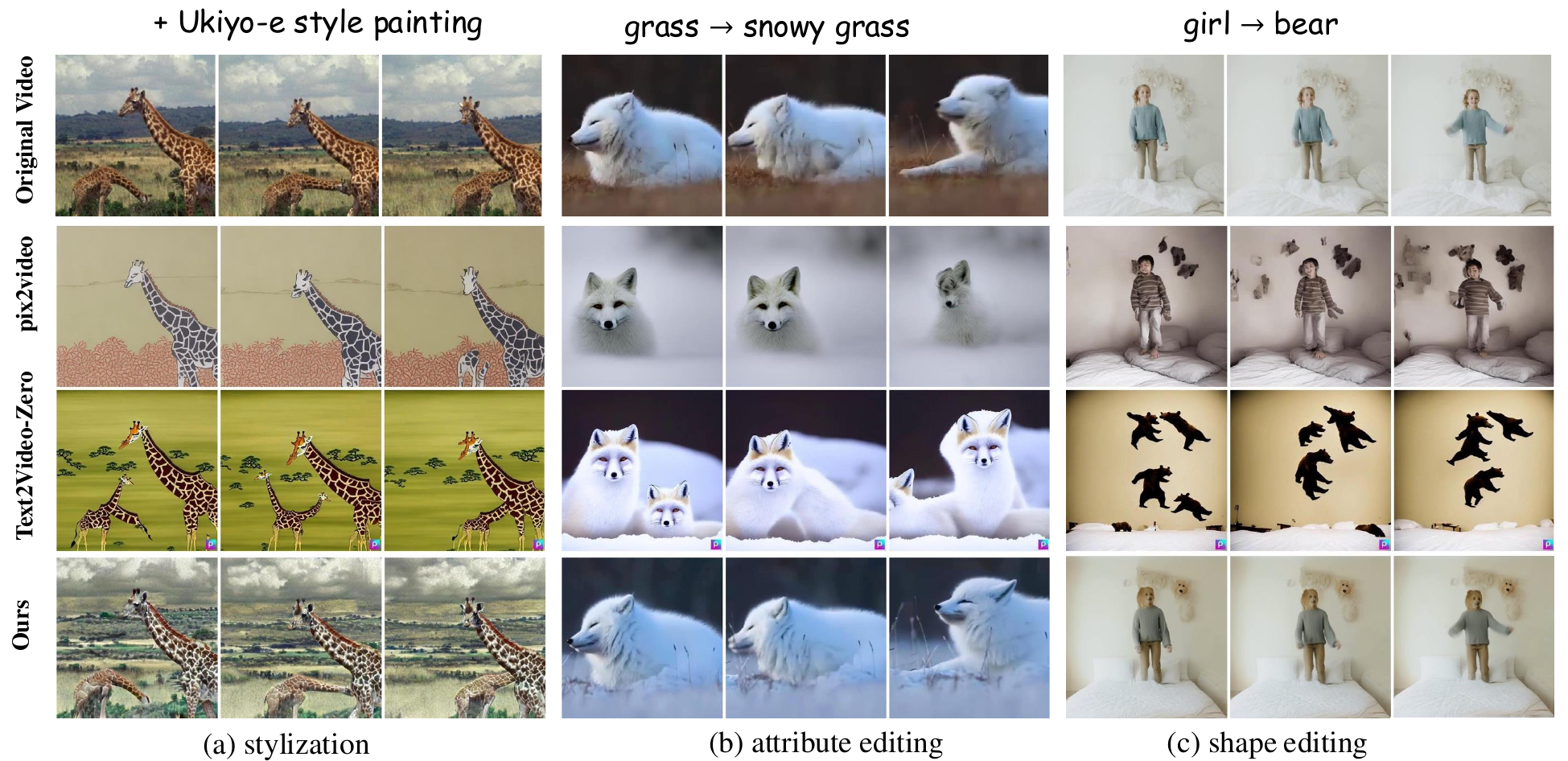} % 替换为您的图片文件名
  \caption{Additional comparative experimental results with several state-of-the-art approaches on three distinct tasks: stylization, attribute editing, and shape editing.} % 图片下方的说明
  \label{fig:compare2} % 为图片定义标签，便于引用
\end{figure*}

\noindent\textbf{Comparision results with other state-of-the-art methods. } We present comparison results excluded from the main text due to space limitations. We also compare FreeMask with Text2Video-Zero~\cite{khachatryan2023text2video} and pix2video~\cite{ceylan2023pix2video}, as shown in Fig~\ref{fig:compare2}. The results show that we perform better than these two tasks regarding overall video quality, prompt alignment, and temporal consistency.

\begin{figure*}[tbp] % 使用 figure* 环境横跨两栏
  \centering % 图片居中
  \includegraphics[width=\textwidth]{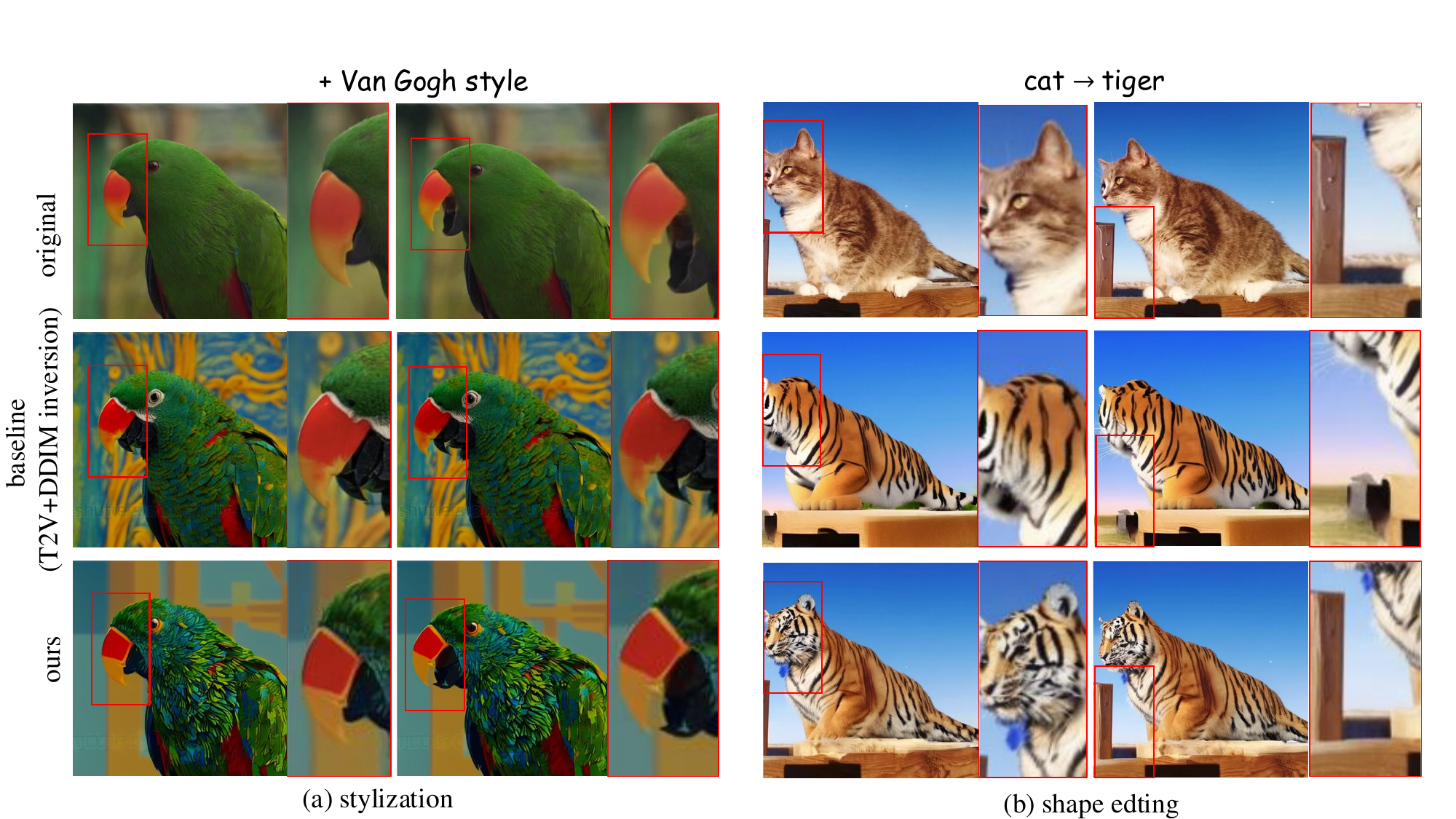} % 替换为您的图片文件名
  \caption{Additional video visualization comparing with baseline.} % 图片下方的说明
  \label{fig:compare_baseline1} % 为图片定义标签，便于引用
\end{figure*}

\noindent\textbf{ Additional video visualization}.  To show the extra benefits of FreeMask compared to the method only based on Text-to-video models and DDIM inversion, we present more video results in Fig~\ref{fig:compare_baseline1}. Using only the T2V model and DDIM inversion gets good temporal consistency, but it often captures only coarse-grained actions and semantic structures, which can lead to suboptimal detail representation.  With our method, both temporal details and semantic nuances are significantly enhanced. For more video examples, please refer to the video file 'FreeMask.mp4' in our supplementary materials.

\subsection{Additional Ablation Results}
\begin{figure}[t] 
  \centering
  \includegraphics[width=\linewidth]{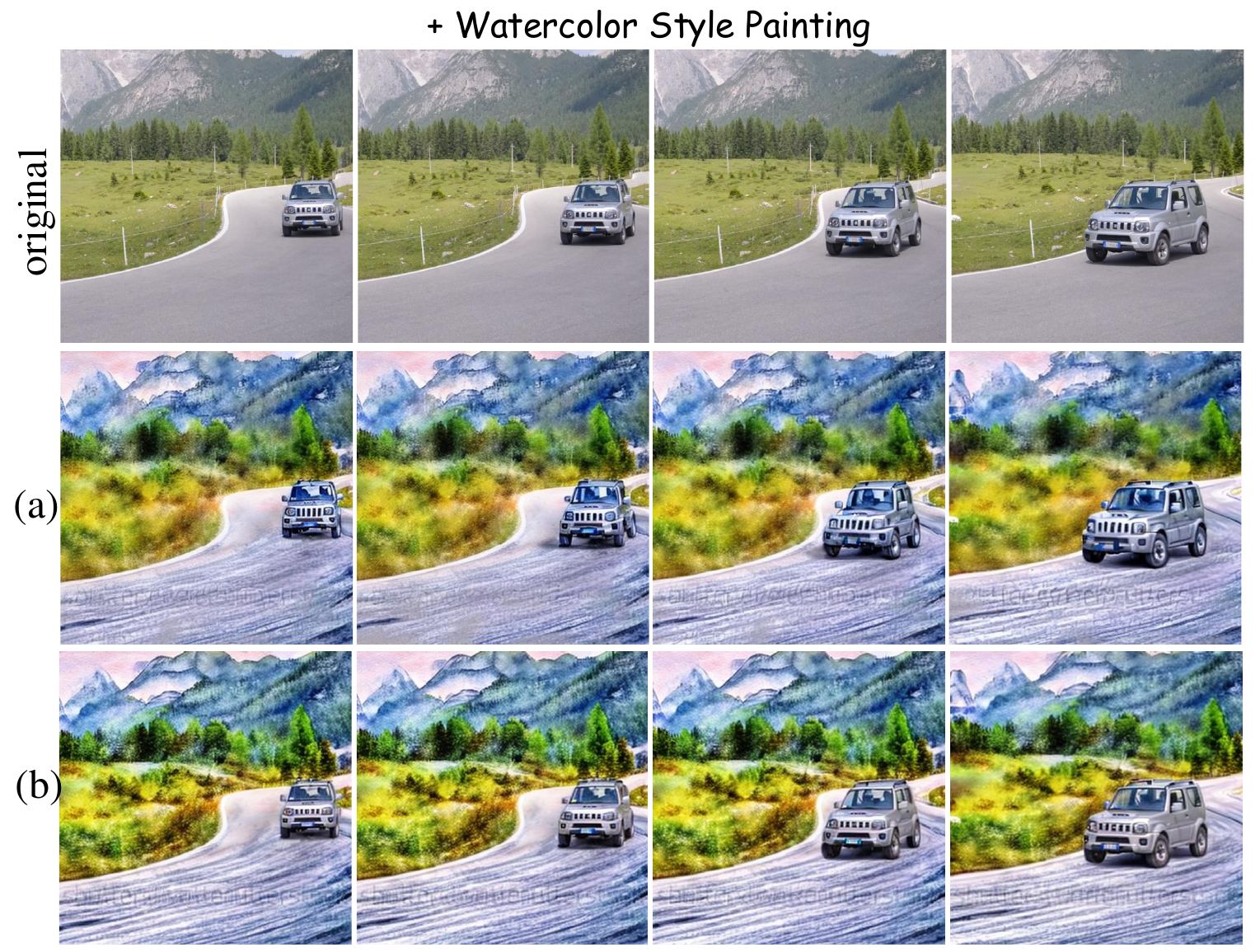} 
  \caption{Ablation results on latent blending ( on Modelscope~\cite{wang2023modelscopet2v}). The prompt is 'Watercolor style painting of a silver jeep driving in the countryside.'  (a) shows stylization with masked attention blending, the quality of dynamic details is improved but not coherent with the original video, and the background appears blurring. (b) shows stylization with masked attention and latent output blending. The detail coherence is much improved, and background details are enriched. 
  }
  \label{fig: ablation4-1} 
\end{figure}

\noindent\textbf{Ablation study of TMMC on stylization. }
We conducted additional ablation experiments on the stylization task to verify the effectiveness of our semantic-adaptive MMC, particularly in how TMMC is used to select masks as shown in Fig~\ref{fig: ablation4-1}, (a) using the same time-aware mask as in shape editing resulted in structural distortion in the background due to mask inaccuracies. In contrast, (b) using a time-agnostic mask significantly reduced the background distortion. These results demonstrate the necessity and effectiveness of the semantic-adaptive MMC in the stylization task.

\noindent\textbf{Ablation study of masked latent fusion with FreeMask. }
Because latent blending is commonly used in video editing and space is limited, we discuss it in this appendix. Here, we apply our MMC-selected mask to the masked fusion of latent outputs of early denoising sampling steps. As is shown in Fig~\ref{fig: ablation4-1}, masked latent blending improves details (such as car headlights in videos), but note that excessive blending may reduce editability, being identical to the original video.

\subsection{Additional Extension Results}
\begin{figure*}[tbp] % 使用 figure* 环境横跨两栏
  \centering % 图片居中
  \includegraphics[width=\linewidth]{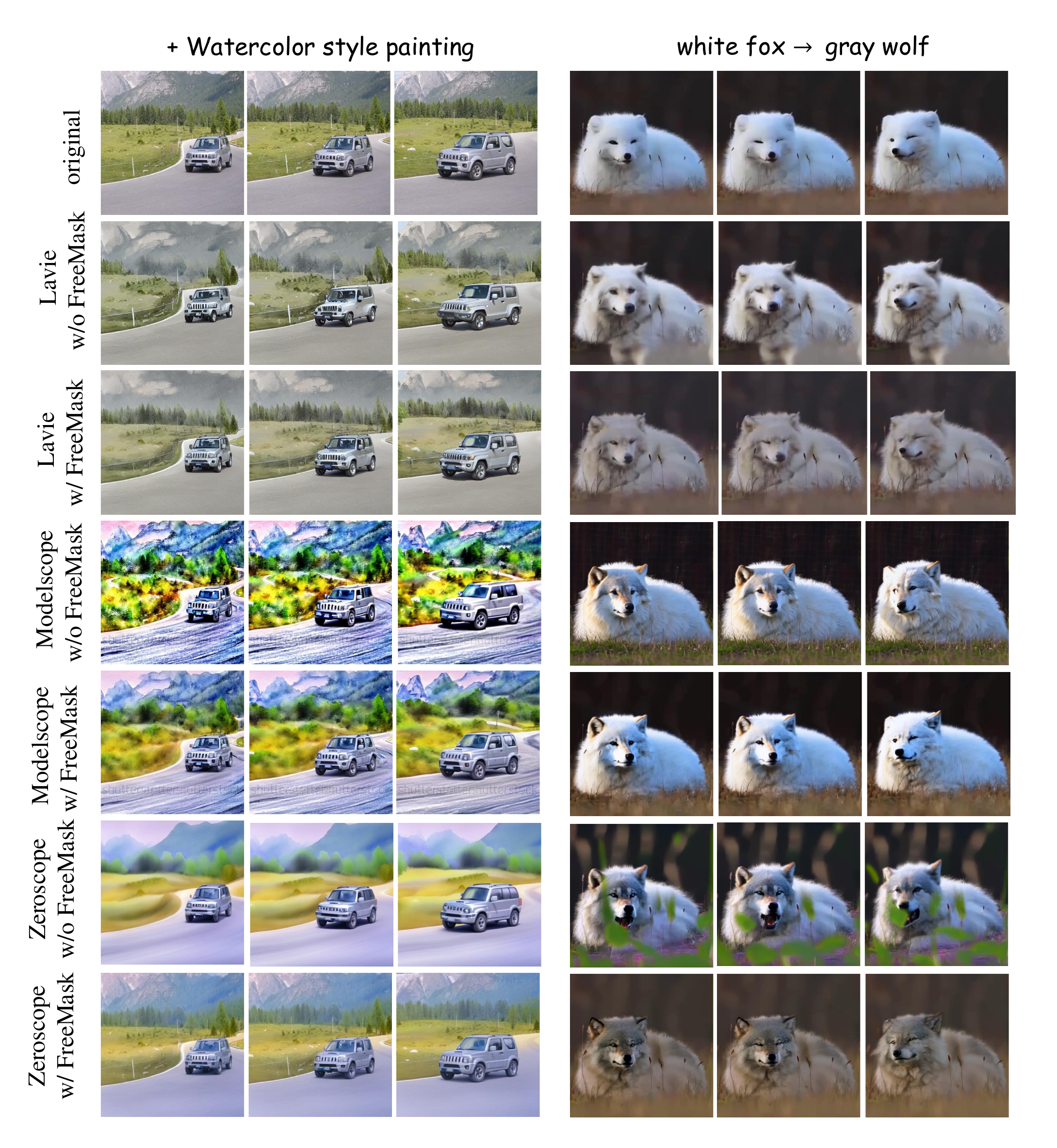} % 替换为您的图片文件名
  \caption{Extension experiments on different T2V models: on Lavie~\cite{wang2023lavie}, Modelscope~\cite{wang2023modelscopet2v} and Zeroscope~\cite{zeroscopev2} specifically, and Zeroscope is mainly adopted in our qualitative and quantitative evaluations.} % 图片下方的说明
  \label{fig:extension2} % 为图片定义标签，便于引用
\end{figure*}

\noindent\textbf{Extension results on other T2V base model.} 
To validate the generalizability of our approach across various text-to-video (T2V) models, we tested it on Lavie, Modelscope, and Zeroscope, all of which extend from image diffusion models with integrated Temporal Layers (including temporal convolutional layers and temporal attention layers). As shown in Fig~\ref{fig:extension2}, despite differences in training strategies and data quality that affect the generative capabilities of these models, our method consistently improves video editing quality, particularly in aligning edited videos with prompts, while enhancing temporal consistency and structural fidelity, demonstrating broad applicability across T2V models.

% \section{Extension Experiments}
% \subsection{Extension on I2V models}
% \subsection{Masked Latent Blending}
% Latent blending is proved effective in Rerender-A-Video~\cite{yang2023rerender} and Text2Video-Zero~\cite{khachatryan2023text2video}, and we can employ our mask guidance on latent blending.

\begin{figure}[t] 
  \centering
  \includegraphics[width=\linewidth]{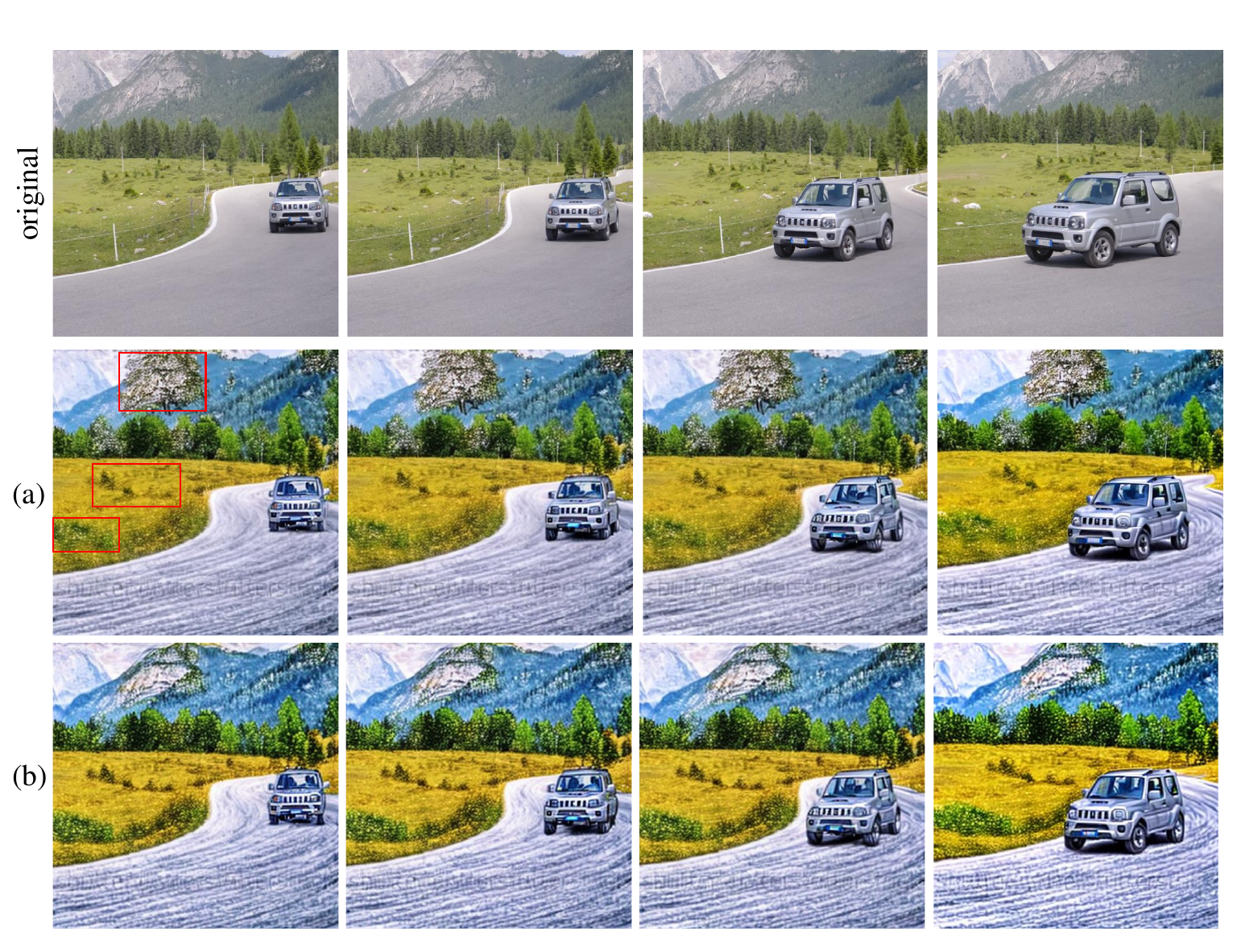} 
  \caption{Ablation study of TMMC on stylization (on Modelscope~\cite{wang2023modelscopet2v}). The prompt is 'Van Gogh style painting of a silver jeep driving in the countryside.'  }
  \label{fig: ablation4-2} 
\end{figure}

% \section{Limitations}
\begin{figure}[t] 
  \centering
  \includegraphics[width=\linewidth]{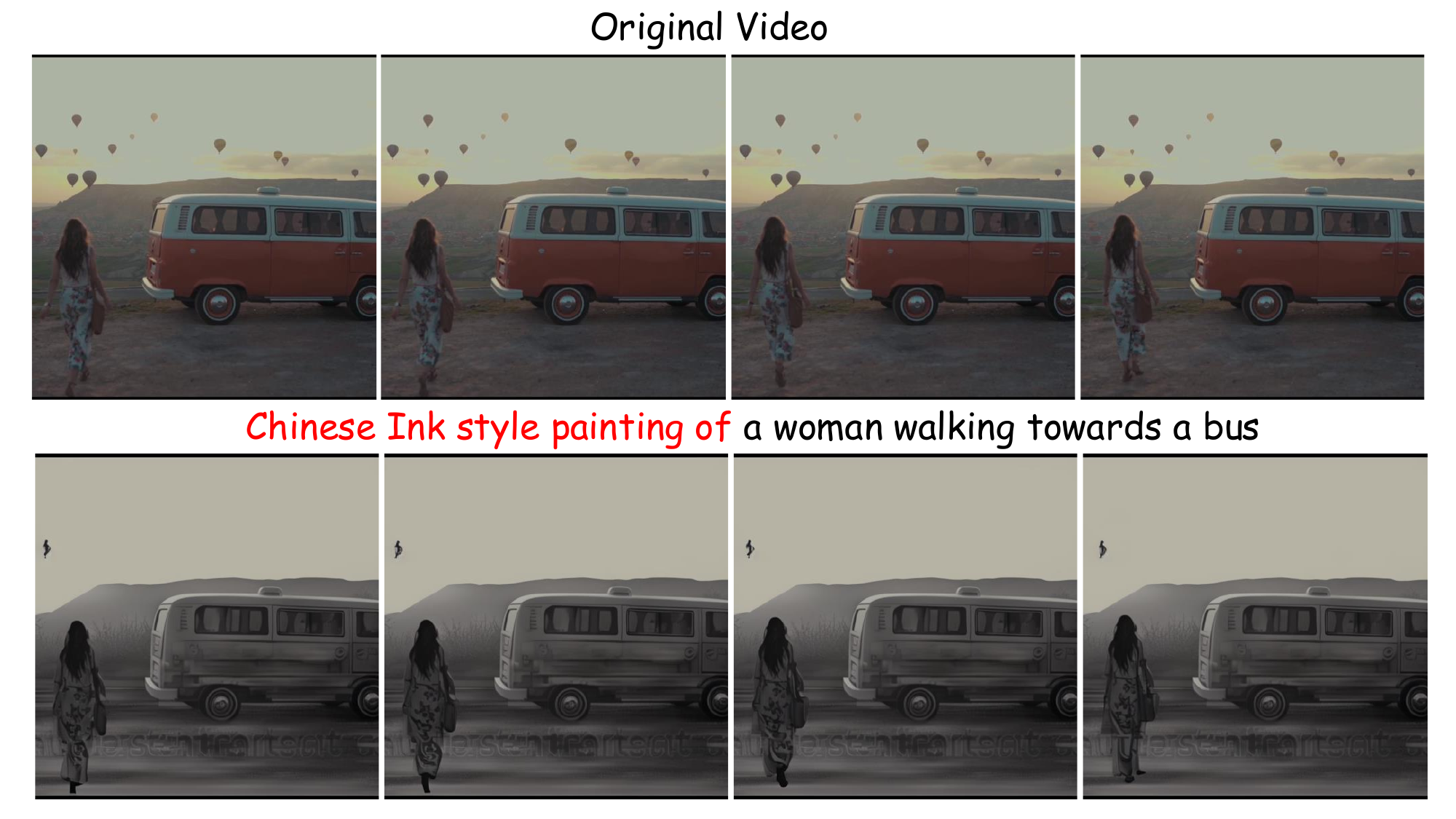} 
  \caption{Limitation. The hot air balloons move to the right in the video clip while a woman walks forward, all captured through a tilt-shift photography technique. In such complex multi-object dynamic scenes, the edited video can only retain information about certain moving objects, such as focusing on the motion of the main subject, the woman, while neglecting background elements. Consequently, some semantic meaning and motion details are lost.}
  \label{fig: limitation} 
\end{figure}
\section{Limitations and Future Work}
% \hangjie{Please also include some future directions for follow-up research efforts.}
Our method encounters challenges when applied to video editing tasks involving multiple objects moving along different trajectories, as shown in Fig~\ref{fig: limitation}. There are two primary reasons for this limitation. First, the base model may not be sufficiently robust in handling complex motion patterns in such videos. Second, beyond mask-based guidance, additional motion information might be necessary to enhance and guide the editing process.

Furthermore, our current approach focuses on appearance editing without addressing motion editing. At present, zero-shot motion editing lacks satisfactory results and often requires the integration of additional motion information for effective guidance. As such, motion editing is not within the scope of our current work.

Addressing these limitations will be the focus of our future research, with particular attention to improving the model's capacity for handling complex motion information and exploring the integration of motion cues to enhance the editing process.

\end{document}